\pdfoutput=1

\documentclass[11pt]{article}

\usepackage[final]{acl}

\usepackage{times}
\usepackage{latexsym}

\usepackage[T1]{fontenc}

\usepackage[utf8]{inputenc}

\usepackage{microtype}

\usepackage{inconsolata}

\usepackage{graphicx}

\usepackage{amsmath}
\usepackage{booktabs}
\usepackage{comment}
\usepackage{mdframed}
\usepackage{array}
\usepackage{enumitem}
\usepackage{adjustbox}
\usepackage{caption}
\usepackage{pdflscape}
\usepackage[utf8]{inputenc}
\usepackage{CJKutf8}
\usepackage{xtab}
\usepackage{url}
\usepackage{graphicx}
\usepackage{subcaption}
\usepackage{xcolor}
\definecolor{deepLavender}{rgb}{0.7, 0.5, 0.8}
\usepackage{tikz} 
\newcommand{\mybox}[4]{
    \begin{figure}[h]
        \centering
    \begin{tikzpicture}
        \node[anchor=text,text width=\columnwidth-0.45cm, draw, rounded corners, line width=1pt, fill=#3, inner sep=2.5mm, align=justify] (big) {\\#4};
        \node[draw, rounded corners, line width=1pt, fill=#2, anchor=west, xshift=5mm] (small) at (big.north west) {#1};
    \end{tikzpicture}
    \end{figure}
}

\newenvironment{hypothesis}%
{%
    \begin{mdframed}[%
        backgroundcolor=gray!20, 
        linewidth=0pt,
        innerleftmargin=10pt,
        innerrightmargin=10pt,
        innertopmargin=10pt,
        innerbottommargin=10pt
    ]%
    \itshape
}%
{%
    \end{mdframed}%
}

\title{Unveiling In-Context Learning: \\A Coordinate System to Understand Its Working Mechanism}

\author{Anhao Zhao \\
  Southwest Jiaotong University \\
  \texttt{zhaoanh@my.swjtu.edu.cn} \\\And
  Fanghua Ye \\
  University College London \\
  \texttt{fanghua.ye.19@ucl.ac.uk} \\\AND 
  Jinlan Fu \\
  National University of Singapore \\
  \texttt{jinlanjonna@gmail.com} \\\And
  Xiaoyu Shen \\
  Digital Twin Institute \\Eastern Institute of Technology, Ningbo \\
  \texttt{xyshen@eitech.edu.cn}
  }

\begin{document}
\maketitle
\begin{abstract}
Large language models (LLMs) exhibit remarkable in-context learning (ICL) capabilities. However, the underlying working mechanism of ICL remains poorly understood. Recent research presents two conflicting views on ICL: One emphasizes the impact of similar examples in the demonstrations, stressing the need for label correctness and more shots. The other attributes it to LLMs' inherent ability of task recognition, deeming label correctness and shot numbers of demonstrations as not crucial. In this work, we provide a \textbf{Two-Dimensional Coordinate System} that unifies both views into a systematic framework. The framework explains the behavior of ICL through two orthogonal variables: {\it whether similar examples are presented in the demonstrations} (\textbf{perception}) and {\it whether LLMs can recognize the task} (\textbf{cognition}). We propose the peak inverse rank metric to detect the task recognition ability of LLMs and study LLMs' reactions to different definitions of similarity. Based on these, we conduct extensive experiments to elucidate how ICL functions across each quadrant on multiple representative classification tasks. 
Finally, we extend our analyses to generation tasks, showing that our coordinate system can also be used to interpret ICL for generation tasks effectively. \footnote{Our code is publicly available at: \url{https://github.com/EIT-NLP/2D-Coordinate-System-for-ICL}.}

\end{abstract}

\section{Introduction}

Large language models (LLMs) have demonstrated impressive in-context learning (ICL) capabilities \citep{brown2020language}, i.e., when provided with few-shot examples, LLMs can effectively perform a broad range of tasks without requiring parameter updates \citep{zhao2021calibrate, min-etal-2022-noisy, su2022welm,wei2023chainofthought}. 
The simplicity of this method, combined with its zero training cost and the versatility of applying a single model across various tasks, has made ICL a promising approach to fully leveraging the potential of LLMs. 

\begin{figure}[t]
\includegraphics[width=\columnwidth]{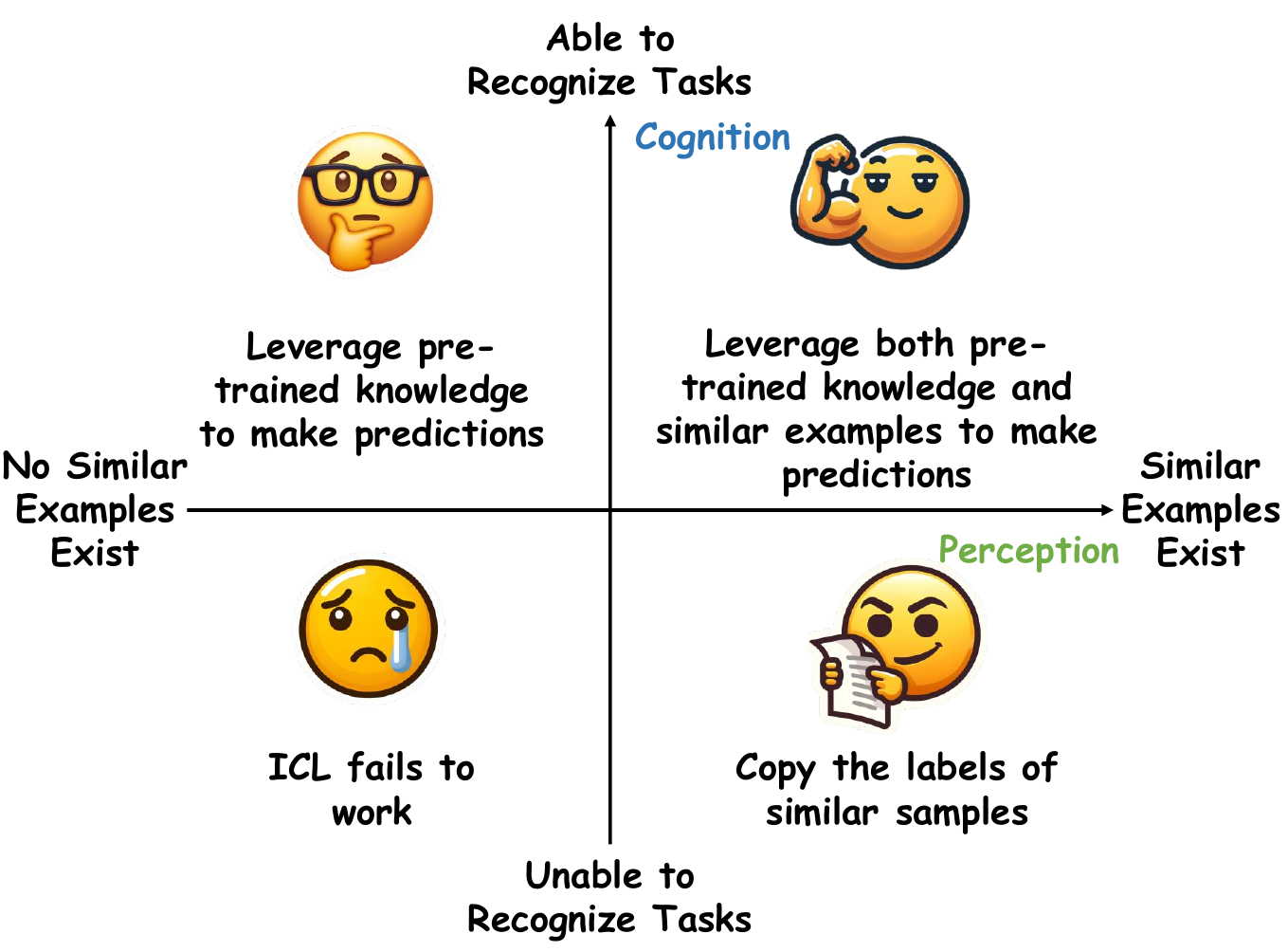}
  \caption{\small An overview of the proposed two-dimensional coordinate system for ICL. The y-axis represents cognition, indicating the model's ability to recognize tasks during ICL, while the x-axis represents perception, reflecting whether similar examples are included in the demonstrations.}

  \label{introduction}
\end{figure}

However, the underlying working mechanism of ICL remains an open question \citep{dai2022can, akyurek2022learning, olsson2022incontextlearninginductionheads, panwar2024incontextlearningbayesianprism}. Existing works hold two conflicting views to explain ICL: The first view argues \emph{LLMs explicitly learn from similar examples in the demonstrations during the inference stage} \cite{liu2021makes}.
Selecting demonstrations similar to the test sample and ensuring their label correctness, or increasing the number of demonstration shots can both improve the performance \cite{rubin-etal-2022-learning, ye2023compositional, levy-etal-2023-diverse,bertsch2024incontext, liu2024unravelingmechanicslearningbaseddemonstration}. The second view, on the contrary, suggests that \emph{LLMs implicitly learn tasks required for downstream applications during the pre-training stage, and in-context demonstrations simply provide cues for them to recognize the task} \cite{xie2022an}.
There have been empirical supports for this hypothesis which show that the performance of ICL is insensitive to label correctness of demonstrations, or the number of demonstration shots~\cite{min2022rethinking,chen2023many,zhang2024impact,zhu2024fine}. 

Although both of the above views hold in specific cases, neither can fully explain the working mechanism of ICL from a holistic perspective. 
In this work, we seek a unified framework encompassing both views to systematically unveil ICL. To do so, we first introduce the peak inverse rank metric to accurately identify the task recognition capability of LLMs. Based on this metric, we observe that LLMs do not always recognize tasks during ICL, even when correct labels and similar examples are provided in the demonstrations. Conversely, successful task recognition also does not necessarily require the presence of similar examples and correct labels.
Hence, we suggest that the effectiveness of ICL should be described through the interactions of these two orthogonal variables, resulting in four distinct ICL scenarios.
To conceptualize this intuitively, we represent each variable along two axes: The x-axis denotes perception, indicating the model's dependence on similar examples for reference. This mirrors human perception, where recognizing similar observed patterns and drawing analogies is crucial for interpreting new information. The y-axis represents cognition, reflecting the model’s task recognition ability. Similar to human cognition, this involves recognizing the right logic learnt from past tasks and reasoning over it to draw the answer, rather than simply replicating observed patterns.
Consequently, all ICL scenarios are visualized within a \textbf{two-dimensional coordinate system}, as depicted in Figure \ref{introduction}.
In each quadrant of the coordinate system, we systematically analyze 8 models spanning up to 40B parameters across three major LLM families. We examine a wide range of classification tasks and have the following main findings:
\begin{itemize}[leftmargin=*]
    \setlength{\itemsep}{0pt} 

    \item In the first quadrant, models are able to recognize the task and similar examples are included in the demonstrations. In this situation, \emph{models can not only leverage their pre-trained knowledge to make predictions but also refer to the labels from similar examples if their pre-trained knowledge is insufficient}. However, if the labels of similar examples are incorrect, smaller models tend to replicate these incorrect labels, while larger models tend to rely on their pre-trained knowledge for making predictions.
    
    \item In the second quadrant, models are able to recognize the task but similar examples are not included in the demonstrations. In this situation, \emph{models primarily leverage their pre-trained knowledge to make predictions}. Moreover, given that each input-label pair plays an identical role in helping models recognize the task, increasing the number of demonstration shots does not significantly enhance the effectiveness of ICL.
    
    \item In the third quadrant, models cannot recognize the task and similar examples are also not included in the demonstrations. In this situation, \emph{ICL fails to work}. Models fail to properly leverage the demonstrations and tend to blindly predict the label of the first example.
    
    \item In the fourth quadrant, models cannot recognize the task but similar examples are included in the demonstrations. In this situation, \emph{models directly replicate the labels of similar examples}. Therefore, the performance of ICL depends entirely on whether the labels of similar examples match the ground truth labels of test samples. Additionally, larger models are better at recognizing similar examples, which increases their tendency to copy the labels from those examples.
\end{itemize}

In general, our findings indicate that for \emph{similar} examples, their label correctness has a \emph{consistent and significant} impact on ICL, especially in scenarios where the model cannot recognize the task and relies heavily on similar examples for inference (below the x-axis). Increasing the number of demonstration shots substantially improves ICL, as it raises the likelihood of matching similar examples. For \emph{dissimilar} examples, label correctness primarily affects the model's \emph{confidence in task recognition} (above the x-axis), but once the task is properly recognized, the effect becomes marginal.

Finally, considering the wide application of ICL in generation tasks \citep{agrawal2022incontext,sia2023incontext,garcia2023unreasonable}, we extend our analyses beyond classification tasks by conducting a thorough case study on a machine translation task. 
This study demonstrates that our coordinate system can also effectively capture the behavior of ICL in generation tasks.
In summary, our proposed coordinate system provides a principled and universal way to understand the working mechanism of ICL.

\section{A 2D Coordinate System for ICL}
\label{Two-Dimensional Coordinate System}

\subsection{The Coordinate Axes}
\label{Two variables in ICL}

\paragraph{Example Similarity.}
\label{Existence of Similar Examples}
Recent works have shown that including similar examples to the test sample in the ICL demonstrations can lead to improved performance outcomes \citep{liu2021makes,rubin-etal-2022-learning,ye2023compositional,levy-etal-2023-diverse,bertsch2024incontext}.
Inspired by this finding, we believe that the presence of examples similar to the test sample in the demonstrations is an important variable for distinguishing different ICL scenarios.

Given the broad concept of similarity, we seek to explore whether the similarity in ICL leans more towards semantic similarity or lexical similarity.
To investigate this question, we perform comprehensive experiments in Appendix \ref{Similarity}, where we construct three different types of examples, including semantically similar but lexically dissimilar examples, lexically similar but semantically opposite examples, and randomly selected examples.
We assign different semantically unrelated label words to these three elements and observe which semantically unrelated word the model predicts.
We find that although ICL tends to slightly favor semantically similar examples over lexically similar ones, the preference for both is significantly greater than randomly selected examples. 
Thus, regardless of whether the similarity is lexical or semantic, as long as demonstrations contain examples of either type, we consider them to contain similar examples.

\paragraph{Task Recognition.}
\label{task recognition}

Instead of capitalizing on similar examples, 
some previous research has demonstrated that in-context demonstrations simply provide information for the model to identify the task to deal with, after which prior knowledge obtained from pretraining data is leveraged to make predictions \citep{xie2022an,min2022rethinking}. This indicates that task recognition is also a crucial factor for ICL. 
However, whether models can always recognize tasks when performing ICL remains an open question.
Hence, there is an urgent need for a method to quantitatively determine the model's task recognition ability.

A recent work by \citet{wang2023label} reveals that label words act as semantic anchors, accumulating information of corresponding demonstrations in the shallow layers. 
The information associated with these anchors is then aggregated in the deeper layers to form the model's final predictions.
Inspired by this, we find that examining whether the hidden states of label tokens at internal layers possess task semantics can serve as an indicator of if ICL has recognized the task.
We use a technique called the logit lens \citep{nostalgebraist2020logit,geva2021transformer,dar2023analyzing}, which projects transformer representations into the vocabulary space, thereby enabling us to interpret abstruse representations in a human-interpretable manner.
Specifically, we project the hidden states of each layer corresponding to the label tokens into the vocabulary space by multiplying them with the pre-trained language modeling head $E$, thus decoding the hidden states of each layer. 
After obtaining the vocabulary distribution of the hidden states for each layer, we calculate the inverse of the rank of the {\it task-representative} token within the vocabulary distribution for each layer. 
We use the peak inverse rank ({\it \textbf{PIR}}) across all layers as our metric to determine whether ICL has recognized the task.
For clarity, we provide the formal definition of {\it \textbf{PIR}} in Appendix \ref{Formal Mathematical Definition of PIR}.
A high \textbf{\textit{PIR}} indicates that ICL has successfully recognized the task, while a low \textbf{\textit{PIR}} suggests a lack of task understanding capabilities.

\begin{figure*}[!t]
   \begin{minipage}[t]{0.32\textwidth}
     \centering
     {\includegraphics[width=\linewidth]{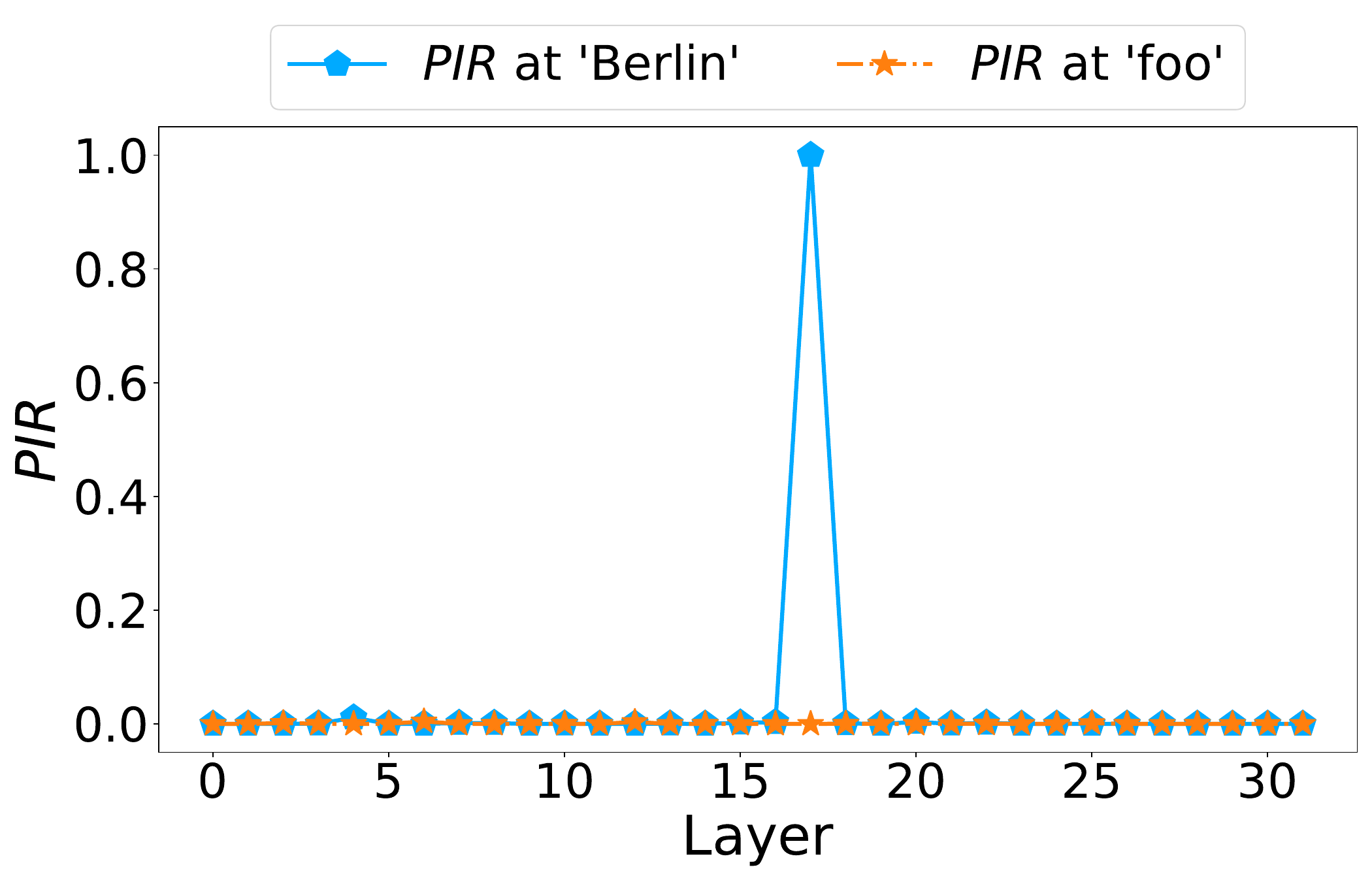}}
     \vspace*{-2.1em}
     \caption{\small The {\it\textbf{PIR}} of {\it "capital"} at the last label token using Llama-2-7B, before and after replacing labels.}
     \label{first_variable}
   \end{minipage}\hfill
   \begin{minipage}[t]{0.32\textwidth}
     \centering
     {\includegraphics[width=\linewidth]{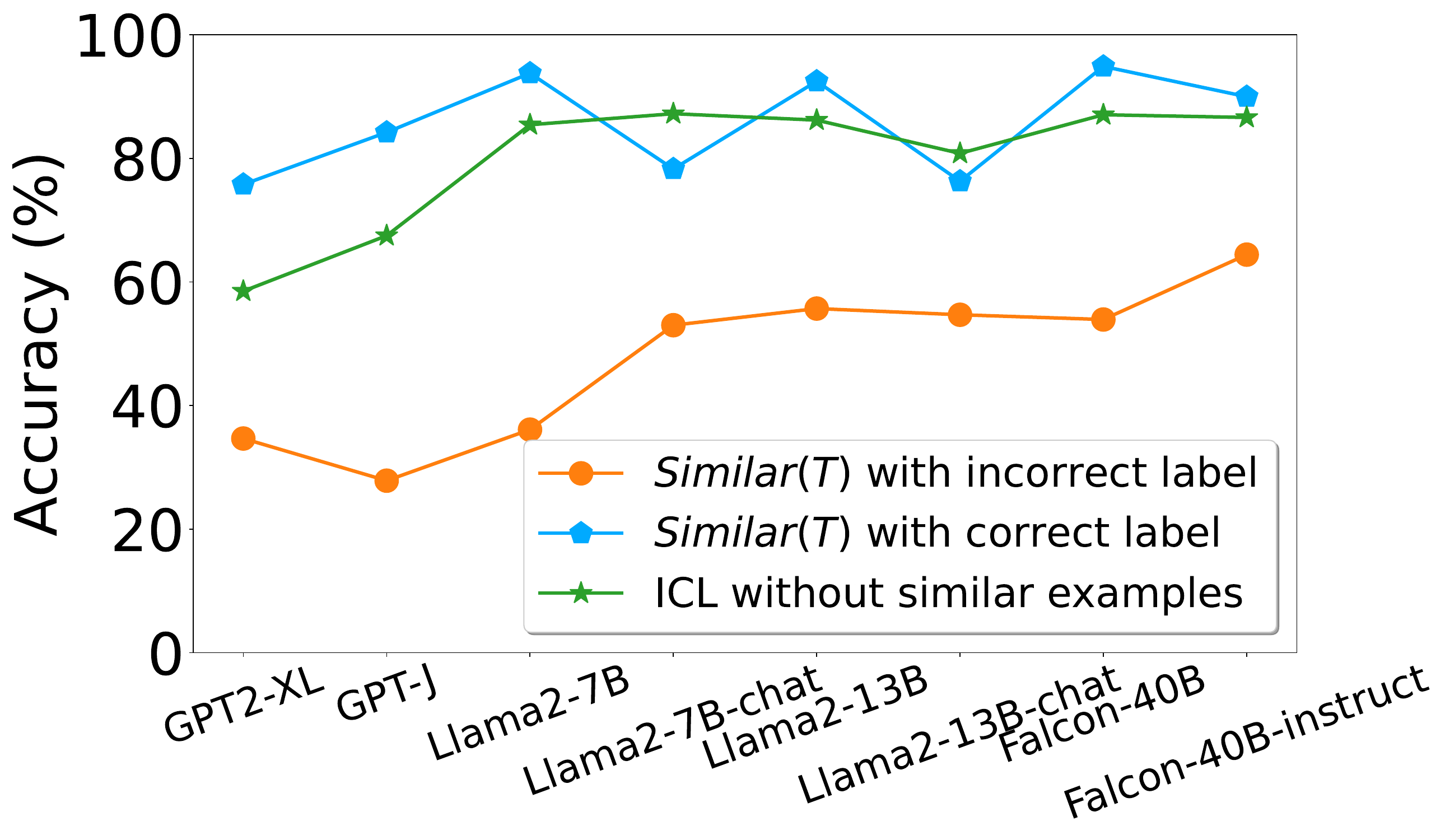}}
     \vspace*{-2em}
     \caption{\small The average ICL accuracy for \texttt{$\text{Similiar}({\text{T}}$}) with correct and incorrect labels, and ICL without similar examples.} 
     \label{first_quadrant}
   \end{minipage}\hfill
   \begin{minipage}[t]{0.32\textwidth}
     \centering
     {\includegraphics[width=\linewidth]{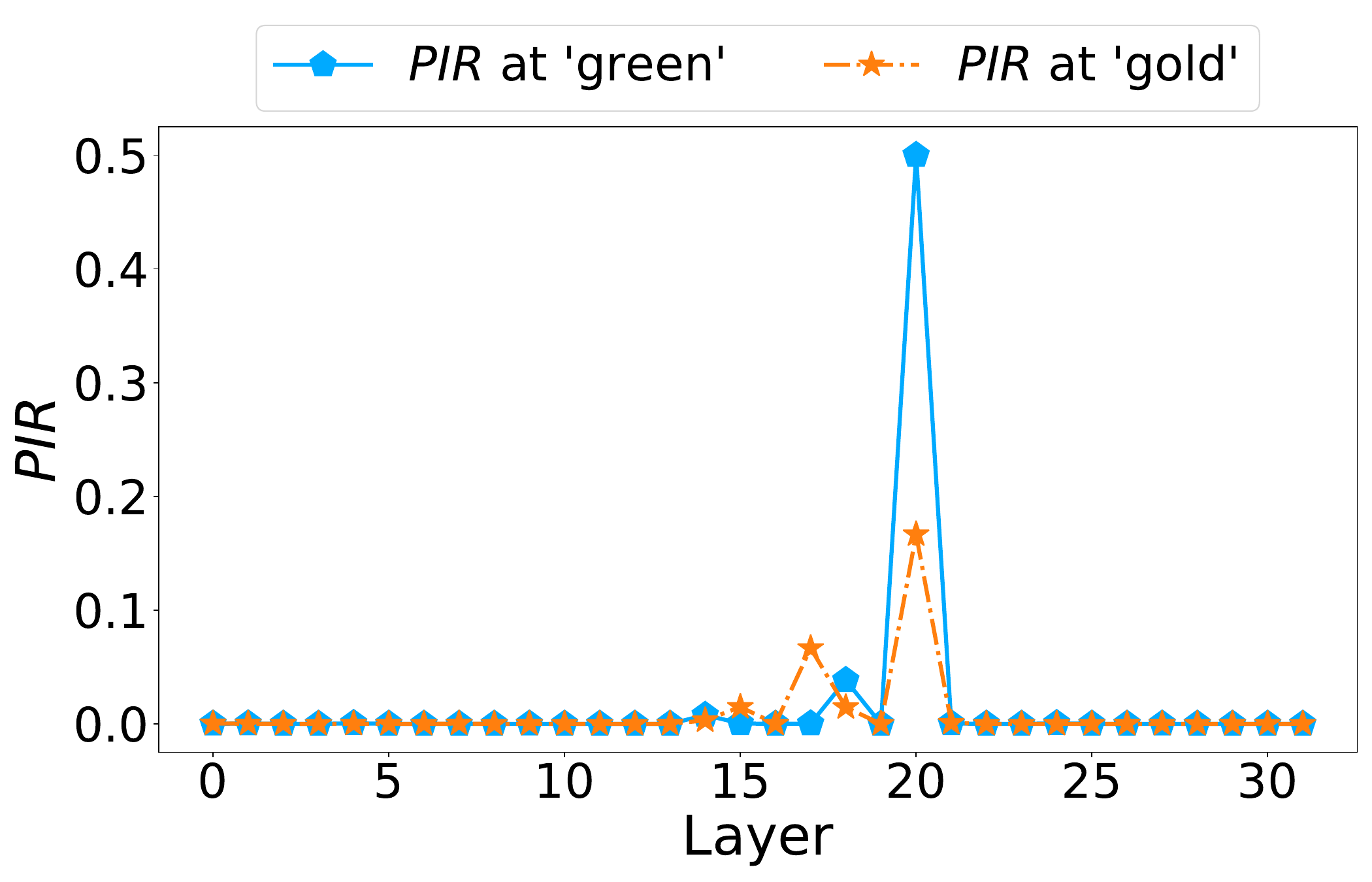}}
     \vspace*{-2em}
     \caption{\small The {\it\textbf{PIR}} of \textit{"color"} at the label token of \texttt{$\text{Similiar}({\text{T}}$}), when the label of \texttt{$\text{Similiar}({\text{T}}$}) is correct and incorrect.}
     \label{first_variable_case_study}
   \end{minipage}
\end{figure*}

To illustrate this, we consider an ICL scenario for the World Capital task: \textit{"{Word: France Label: Paris Word: Germany Label: Berlin Word: Italy Label:}"}.
We select the last label word \textit{"{Berlin}"} to report the \textbf{\textit{PIR}} of the task-representative token \textit{"{capital}"} in Figure \ref{first_variable}, where the \textbf{\textit{PIR}} reaches 1 at layer 17.
For the same task, we replace all labels with semantically irrelevant words to prevent ICL from recognizing the task. 
This setting, called {\bf task learning}, was first introduced by \citet{pan2023context}. 
Concretely, we replace the first label \textit{"{Paris}"} with \textit{"{bar}"} and the second label \textit{"{Berlin}"} with \textit{"{foo}"}. 
We still select the last label \textit{"{foo}"} to report the \textbf{\textit{PIR}} of \textit{"{capital}"} in Figure \ref{first_variable}, where the \textbf{\textit{PIR}} drops to 0.

Based on \textit{\textbf{PIR}}, we observe that models do not always recognize tasks during ICL, even when the demonstrations are entirely composed of correct input-label pairs (refer to Appendix \ref{Detailed Proof of Models" Task Recognition on Classification Datasets}). 
Furthermore, we demonstrate that the presence of similar examples and the ability of task recognition are orthogonal to each other (refer to Appendix \ref{Detailed Proof: The Orthogonality of Similar Examples Presence and LLMs' Task Recognition Ability}).
Given this, we take whether models can recognize tasks during the execution of ICL as the second variable to distinguish different ICL scenarios.
We will use \textbf{\textit{PIR}} as the criterion to select datasets for ICL that can and cannot recognize tasks.

\subsection{ICL Scenario Exploration}
\label{Scenario Exploration}
Following the above discussions, ICL scenarios can be described comprehensively using two variables: task recognition and example similarity.
These two variables result in four combinations, which can be visualized in a two-dimensional coordinate system (see Figure \ref{introduction}). 
In the positive half of the y-axis, the model is capable of recognizing tasks, while in the negative half, it is not.
Similarly, the positive half of the x-axis signifies the presence of similar examples, whereas their absence is indicated in the negative half.

Importantly, we establish this two-dimensional coordinate system not only to conceptualize the four possible combinations of ICL scenarios but also to consider the two variables on a continuous scale. 
Specifically, as the model's confidence in recognizing tasks increases, denoted by \textbf{\textit{PIR}} approaching one, the y-coordinate value rises. 
Likewise, as the similarity between the provided examples and the test sample increases, the x-coordinate value rises.
In the following section, we will provide a detailed description of how ICL works within each quadrant of the coordinate system.

\section{Experiments and Results}
\subsection{Experimental Settings}
Given that it is not intuitively clear whether models can recognize the task in a particular dataset, which in fact must be verified using the metric \textbf{\textit{PIR}}, we directly enumerate the classification datasets in which models can and cannot recognize the task. 
For detailed proofs, please refer to Appendix \ref{Detailed Proof of Models" Task Recognition on Classification Datasets}.

\paragraph{Datasets in Which Models Can Recognize Tasks.}
These datasets are used for the upper part of the x-axis. We employ the Stanford Sentiment Treebank Binary ({\bf SST-2}) \citep{socher-etal-2013-recursive} for sentiment analysis. In addition, 
we create two datasets for the \textbf{World Capitals} and \textbf{Reasoning about Colored Objects} tasks, which contain 50 hand-crafted pairs of {\it country-capital} and {\it object-color}, respectively. The detailed data are provided in Appendix \ref{Detailed Data for World Capitals and Colored Objects Tasks}.

\paragraph{Datasets in Which Models Cannot Recognize Tasks.}
These datasets are used for the lower part of the x-axis. We utilize the Text REtrieval Conference ({\bf TREC}) Question Classification dataset \citep{li2002learning, hovy2001toward} for question type classification and the EmoContext ({\bf emo}) \citep{chatterjee-etal-2019-semeval} for emotion classification.

\paragraph{Models.} 
We adopt a comprehensive suite of models, including GPT2-XL (1.61B) \citep{radford2019language} and GPT-J (6B) \citep{mesh-transformer-jax} from the GPT series; Llama-2-7B, Llama-2-13B, and their instruction-tuned counterparts from the Llama series \citep{touvron2023llama}; and Falcon-40B, along with its instruction-tuned variant from the Falcon series \citep{falcon40b}. 


\paragraph{Prompt Format.}  
We use neutral delimiters to avoid providing task-specific information. See Appendix \ref{Delimiters Used for Each Dataset} for details.

\paragraph{Accuracy Metric.}
We consider an individual prediction correct only when the token with the highest logit within the {\it entire} vocabulary matches the label of the test sample.
Accuracy is the proportion of correct predictions across the whole dataset.

\begin{figure*}[!t]
   \begin{minipage}[t]{0.32\textwidth}
     \centering
     {\includegraphics[width=\linewidth]{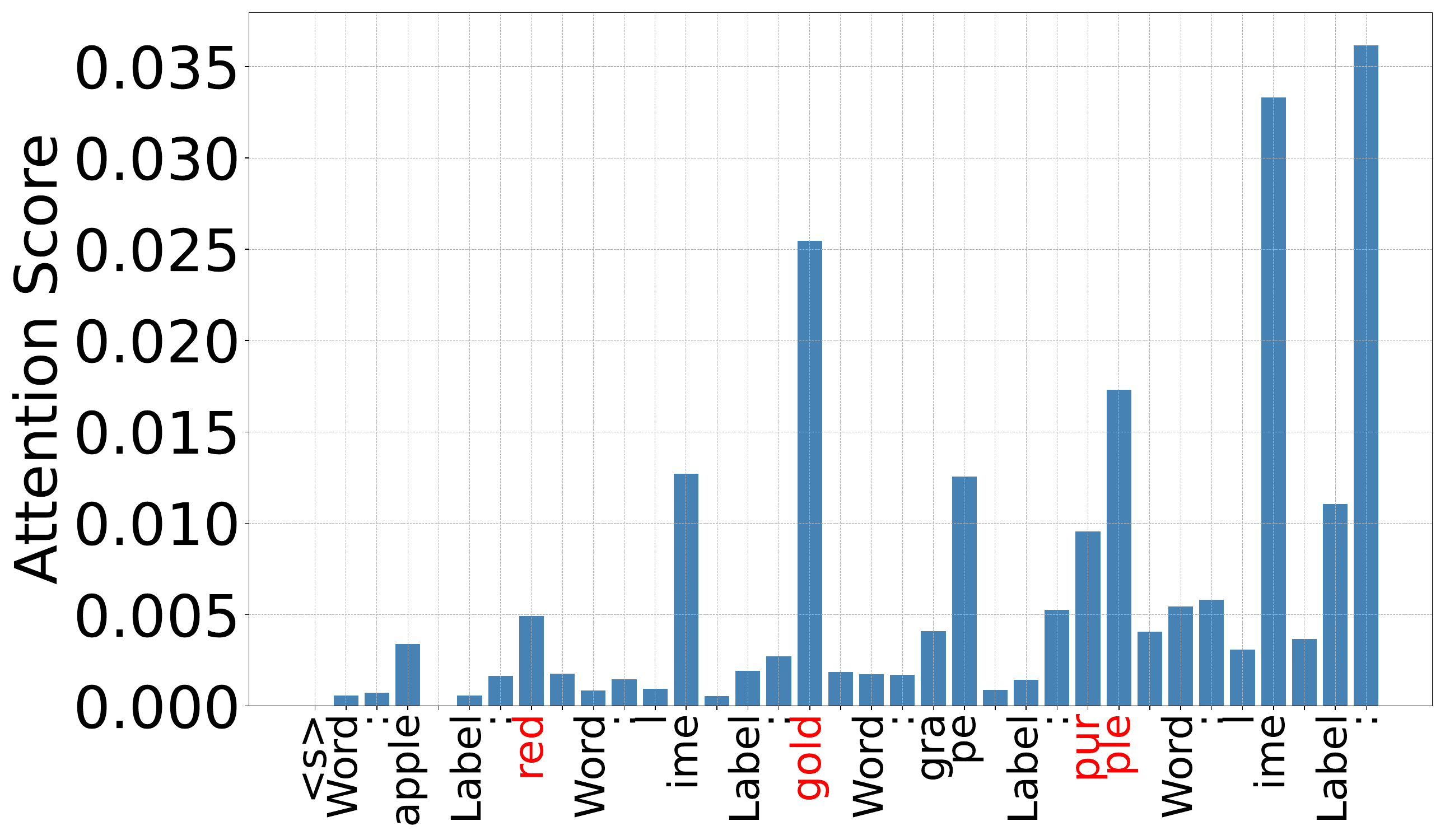}}
     \vspace*{-2em}
     \caption{\small In the 20th layer, the attention scores of the last token ":" for all tokens. All label tokens are marked in red.} 
     \label{attention_layer20}
   \end{minipage}\hfill
   \begin{minipage}[t]{0.32\textwidth}
     \centering
     {\includegraphics[width=\linewidth]{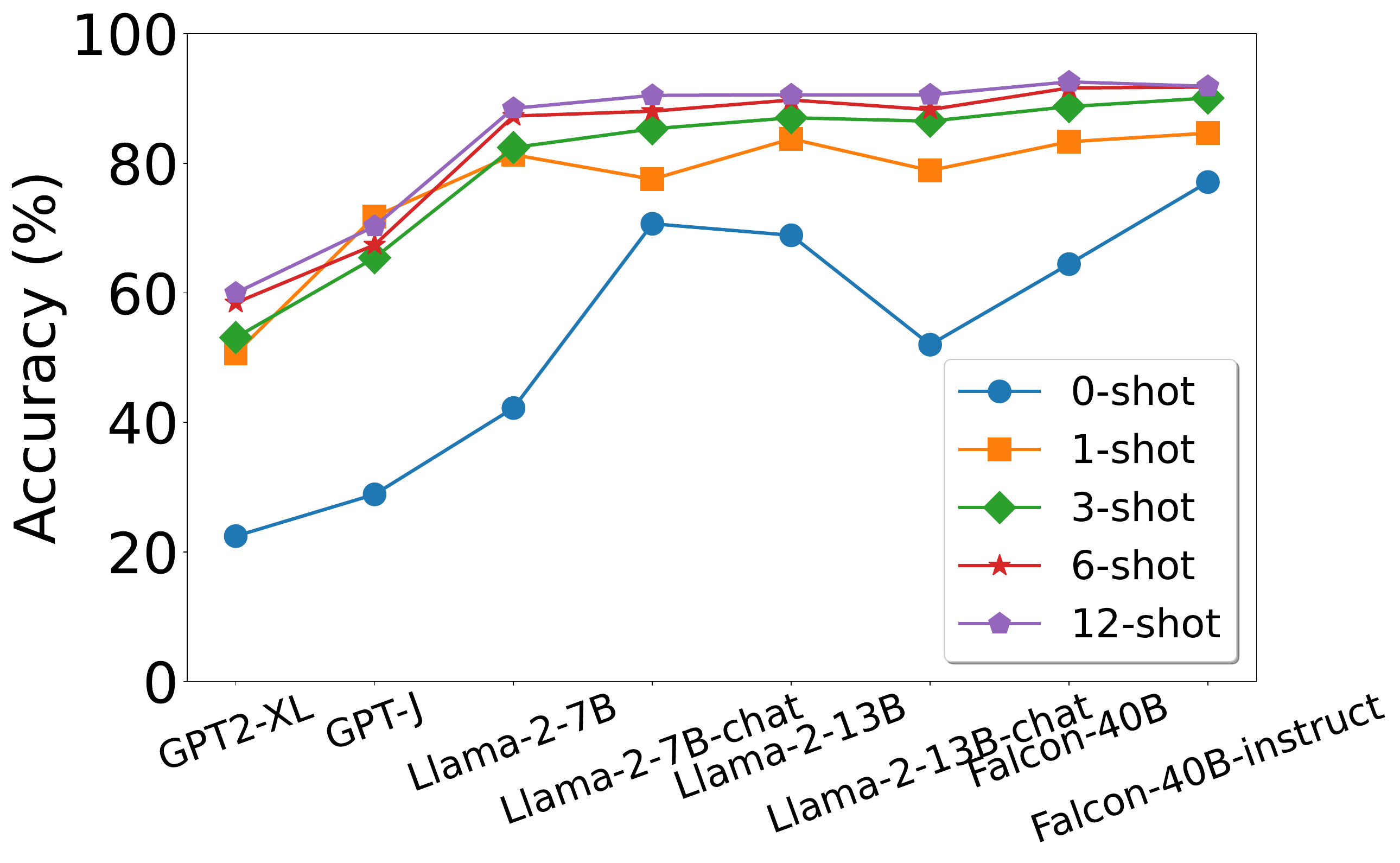}}
     \vspace*{-2em}
     \caption{\small Averaged accuracy of zero-shot and ICL with varying numbers of demonstrations in the second quadrant.} 
     \label{second_quadrant}
   \end{minipage}\hfill
   \begin{minipage}[t]{0.32\textwidth}
     \centering
     {\includegraphics[width=\linewidth]{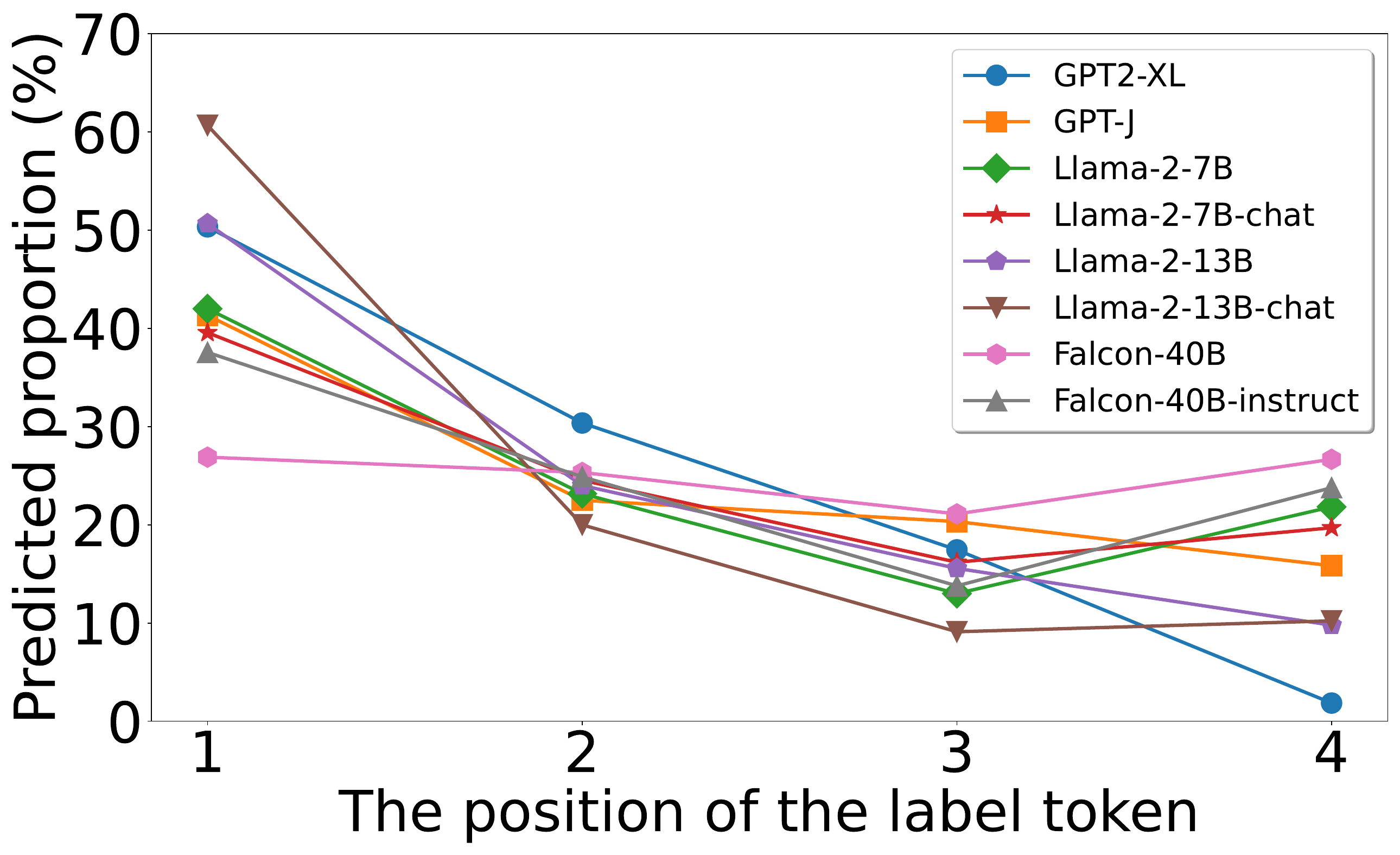}}
     \vspace*{-2em}
     \caption{\small For the emo dataset, the preference of different models for label tokens at different absolute positions.} 
     \label{position_bias}
   \end{minipage}
\end{figure*}

\subsection{First Quadrant}
\label{First Quadrant}
In this quadrant, \textbf{models can recognize tasks when performing ICL, and the demonstrations contain examples similar to the test sample.}
In this situation, since models can rely not only on their pre-trained knowledge after task recognition but also on the labels of similar examples to make predictions, we aim to determine how these two factors work together.

\paragraph{Implementation Details.}
Here, we consider an extreme setup where we artificially add the test sample into the demonstration, identifying it as a similar example with the highest similarity, denoted as \texttt{$\text{Similiar}({\text{T}}$}).
We consider three settings: \texttt{$\text{Similiar}({\text{T}}$}) with an incorrect label, \texttt{$\text{Similiar}({\text{T}}$}) with the correct label, and ICL without similar examples. 
For the first setting, we select a label from the label space that differs from the test sample's label and assign it to the \texttt{$\text{Similiar}({\text{T}}$}).
For the second setting, we select the label from the label space that matches the test sample's label and assign it to the \texttt{$\text{Similiar}({\text{T}}$}). 
For the third setting, we select examples from the training dataset with no similarity to the test sample to serve as demonstrations.
We use $k=6$ in-context examples. 
The results reflect averages from five random seeds and all datasets in which models can recognize tasks.


\paragraph{Experimental Results.}
Figure \ref{first_quadrant} shows that
(1) adding \texttt{$\text{Similiar}({\text{T}}$}) with the correct label in the demonstrations generally improves performance compared to ICL without similar examples (i.e., ICL falling within the second quadrant), although there is a slight performance decline for Llama-2-7B-chat and Llama-2-13B-chat.
This indicates that models can not only utilize their pre-trained knowledge but also refer to the correct label of \texttt{$\text{Similiar}({\text{T}}$}) when their pre-trained knowledge is insufficient.
(2) However, including \texttt{$\text{Similiar}({\text{T}}$}) with an incorrect label in the demonstrations induces a state of "confusion", making models neither completely rely on their pre-trained knowledge for predictions (performing worse than ICL without similar examples) nor completely overwrite their predictions with the label of \texttt{$\text{Similiar}({\text{T}}$}).



\paragraph{In-Depth Analysis.}
To comprehend the causes of "confusion" experienced by models, we present insights through a case study on the Reasoning about Colored Objects task using Llama-2-7B.
We first consider the scenario when the label of \texttt{$\text{Similiar}({\text{T}}$}) is correct: \textit{"{Word: apple Label: red Word: lime Label: green Word: grape Label: purple Word: lime Label:}"}. 
We report the \textbf{\textit{PIR}} of the task-representative token \textit{"{color}"} at the label \textit{"{green}"} of \textit{"{lime}"}, as shown in Figure \ref{first_variable_case_study}. 
Next, we substitute the label token \textit{"{green}"} with the incorrect color \textit{"{gold}"}, thereby constructing a \texttt{$\text{Similiar}({\text{T}}$}) with an incorrect label in the demonstrations.
Similarly, we report the \textbf{\textit{PIR}} of the task-representative token \textit{"{color}"} at the label \textit{"{gold}"} of \textit{"{lime}"}, also depicted in Figure \ref{first_variable_case_study}. 
As illustrated, replacing the correct label with an incorrect one decreases the model's confidence in the Reasoning about Colored Objects task at the label token of \texttt{$\text{Similiar}({\text{T}}$}).
Additionally, due to \texttt{$\text{Similiar}({\text{T}}$}) having the highest semantic and lexical similarity, the last token of the input at the intermediate layer assigns the highest attention score to \textit{"{gold}"} among all label tokens.
For instance, consider the 20th layer, as shown in Figure \ref{attention_layer20}.
Consequently, at this layer, the hidden state for \textit{"{gold}"} contributes more significantly to the residual stream of the last token compared to the hidden states of other label tokens. 
This leads to a reduction of task semantics and an increase of the semantics associated with the word \textit{"{gold}"} within the hidden states of the last token. 
As a result, the model faces uncertainty about whether to rely on pre-trained knowledge for making predictions or to directly output the \textit{"{gold}"} token, leading to the "confusion" phenomenon.

Additionally, as observed in Figure \ref{first_quadrant}, when encountering the "confusion" phenomenon, models with smaller parameter sizes tend to output incorrect labels, whereas models with larger parameter sizes are more likely to rely on their pre-trained knowledge for the output.
This indicates that when the label of \texttt{$\text{Similiar}({\text{T}}$}) is incorrect, the confidence in the task at that label token increases as the model size increases.


\mybox{{\bf Conclusion}}{gray!40}{gray!10}{In the first quadrant, models can leverage their pre-trained knowledge to make predictions once they recognize the task and can also refer to the labels from similar examples if their pre-trained knowledge is insufficient. However, if the labels of similar examples are incorrect, smaller models tend to replicate these incorrect labels, while larger models tend to rely on their pre-trained knowledge for making predictions.}

\subsection{Second Quadrant}
\label{Second Quadrant}
In this quadrant, \textbf{models can recognize tasks when performing ICL, but the demonstrations do not contain examples similar to the test sample.}
Due to the absence of similar examples, the phenomenon described in the first quadrant—where the incorrect label semantics of similar examples significantly affect the hidden states of the last token—does not occur. 
Therefore, randomly replacing labels in this quadrant does not substantially impact ICL performance, which is consistent with existing work \citep{min2022rethinking}. 
The detailed experimental results can be found in Appendix \ref{Impact of Random Label Replacement in Second Quadrant}.

Here, we would like to discuss a new question: 
\textit{Since the label token of each example provides the same task semantics (because all examples are dissimilar to the test sample), does this imply that we can achieve good ICL performance with only a very small number of examples?}

\paragraph{Implementation Details.}
We use the "zero-shot" approach as our baseline, where the model is only given an instruction that specifies the task. 
Detailed instructions for each dataset can be found in Appendix \ref{Detailed instructions for each dataset}. 
Subsequently, we remove the instruction and provide the model with only demonstrations, then incrementally increase the number of shots.  
The results are averaged over five random seeds and datasets in which models can recognize tasks.

\paragraph{Experimental Results.}
The experimental results are shown in Figure \ref{second_quadrant}.
It can be observed that with only a single input-label pair, the performance of ICL significantly surpasses that of the zero-shot setting. 
Further increasing the number of demonstration shots results in very limited performance improvement. 
These results confirm our hypothesis: The roles of each label token overlap, and adding more examples merely reinforces the model's confidence in correctly identifying the task.

\mybox{{\bf Conclusion}}{gray!40}{gray!10}{ In the second quadrant, models primarily leverage their pre-trained knowledge to make predictions. Moreover, given that each input-label pair plays an identical role in helping models recognize tasks, increasing the number of in-context examples does not significantly enhance the effectiveness of ICL.}

\subsection{Third Quadrant}
\label{Third Quadrant}
In this quadrant, \textbf{models cannot recognize tasks when performing ICL, and the demonstrations also do not contain examples similar to the test sample.}
This represents the worst-case scenario among all quadrants.
As illustrated in Figure \ref{third_quadrant}, the performance of one-shot ICL in this quadrant is significantly worse than the zero-shot setting for all models.
In this quadrant, what do the models rely on to make predictions when performing ICL?

\paragraph{Implementation Details.}
We consider an ICL setting where each example's label corresponds to a different label class from the dataset, covering all labels of the dataset (i.e., 4-shot ICL for emo and 6-shot ICL for TREC), to observe the predictive behavior of ICL.
To ensure that the ICL output remains within the label space, we prefix the input with instructions to limit the output range without specifying the task.
We select an equal number of samples from each label class in the dataset to serve as the set of test samples.
For each test sample, we manually select demonstration examples that have virtually no semantic similarity or lexical similarity.
Instead of focusing on whether the output matches the ground truth, we focus on the absolute position, specifically which position's label token will be predicted. 
Here we report the experimental results on the emo dataset in Figure \ref{position_bias}. The results for TREC can be found in Appendix \ref{Positional Bias of TREC Dataset in the Third Quadrant}.

\paragraph{Experimental Results.}
As shown in Figure \ref{position_bias}, we find that models exhibit a strong positional bias in this quadrant. 
Specifically, we observe that for almost all models, there is a significantly high proportion of instances for which the label of the first input-label pair is predicted. 
In contrast, the proportion of predictions for the labels of other pairs is notably lower. 
We attribute this to the attention sink phenomenon discovered by \citet{xiao2024efficient}, where models tend to allocate more attention to the initial tokens during prediction.


\mybox{{\bf Conclusion}}{gray!40}{gray!10}{ In the third quadrant, ICL fails to work. Specifically, models fail to leverage the ICL content for making predictions and tend to predict the label of the first example.}

\subsection{Fourth Quadrant}
\label{Fourth Quadrant}
In this quadrant, \textbf{models cannot recognize tasks when performing ICL, but the demonstrations contain examples similar to the test sample.} 
Since models in this context can only reference similar examples, we hypothesize that the accuracy of ICL predictions hinges on whether the labels of these similar examples align with the ground-truth label of the test sample.

\paragraph{Implementation Details.}
Similar to the experimental setup in the first quadrant, during each ICL inference, we randomly select a label from the dataset that differs from the test sample's label and assign it to \texttt{$\text{Similiar}({\text{T}}$}).
Since we aim to verify whether models in this quadrant rely on the labels of similar examples to make predictions, we use the proportion of predictions for the incorrect label assigned to \texttt{$\text{Similiar}({\text{T}}$}) as our accuracy metric.
We adopt $k=12$ in-context examples.  
The results reflect averages from five random seeds and all datasets in which models cannot recognize tasks.

\begin{figure*}[!t]
   \begin{minipage}[t]{0.32\textwidth}
     \centering
     {\includegraphics[width=\linewidth]{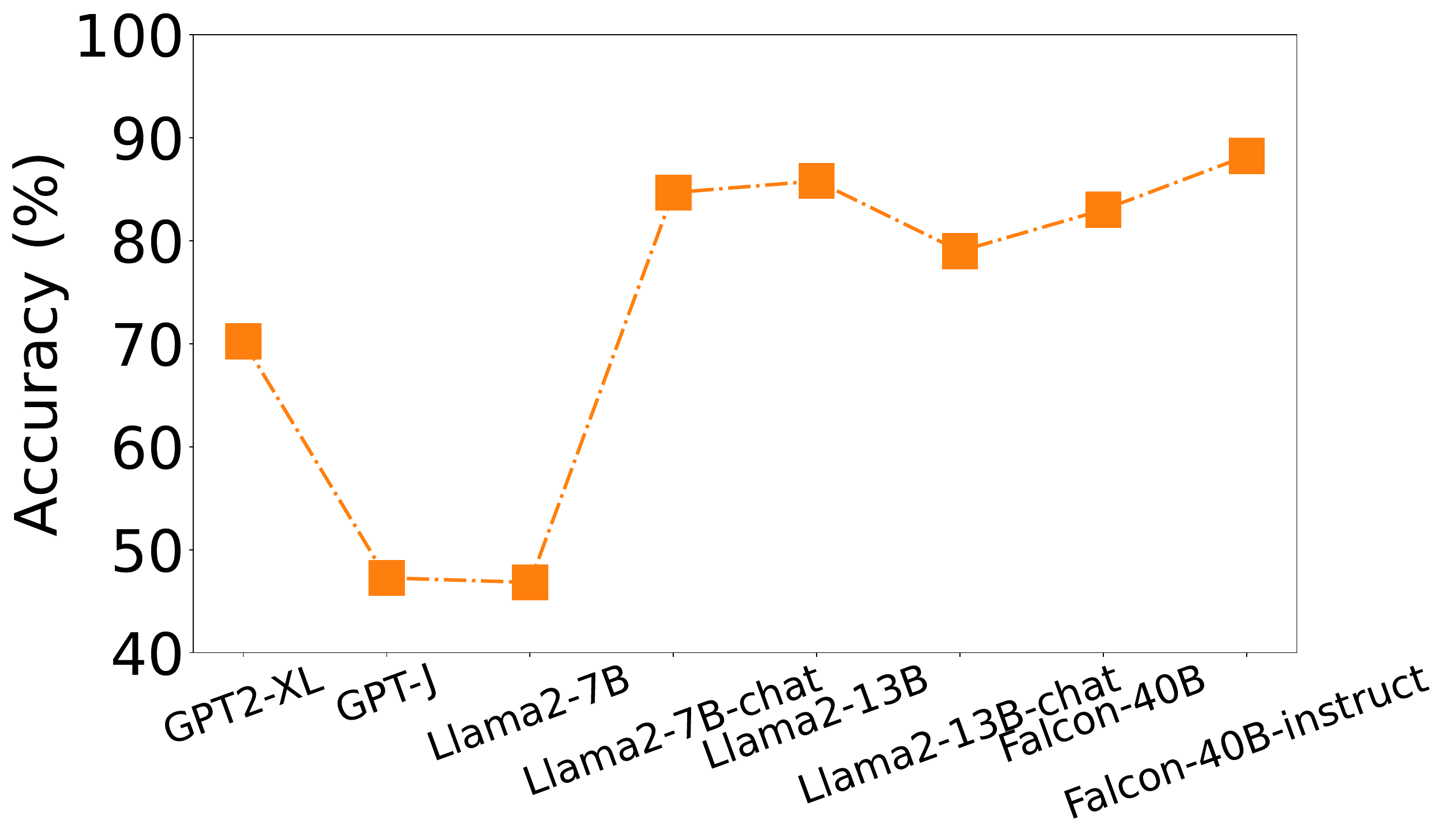}}
     \vspace*{-2em}
     \caption{\small The proportion of predictions for the incorrect label corresponding to \texttt{$\text{Similiar}({\text{T}}$}) on various models.}
     \label{fourth_quadrant}
   \end{minipage}\hfill
   \begin{minipage}[t]{0.32\textwidth}
     \centering
     {\includegraphics[width=\linewidth]{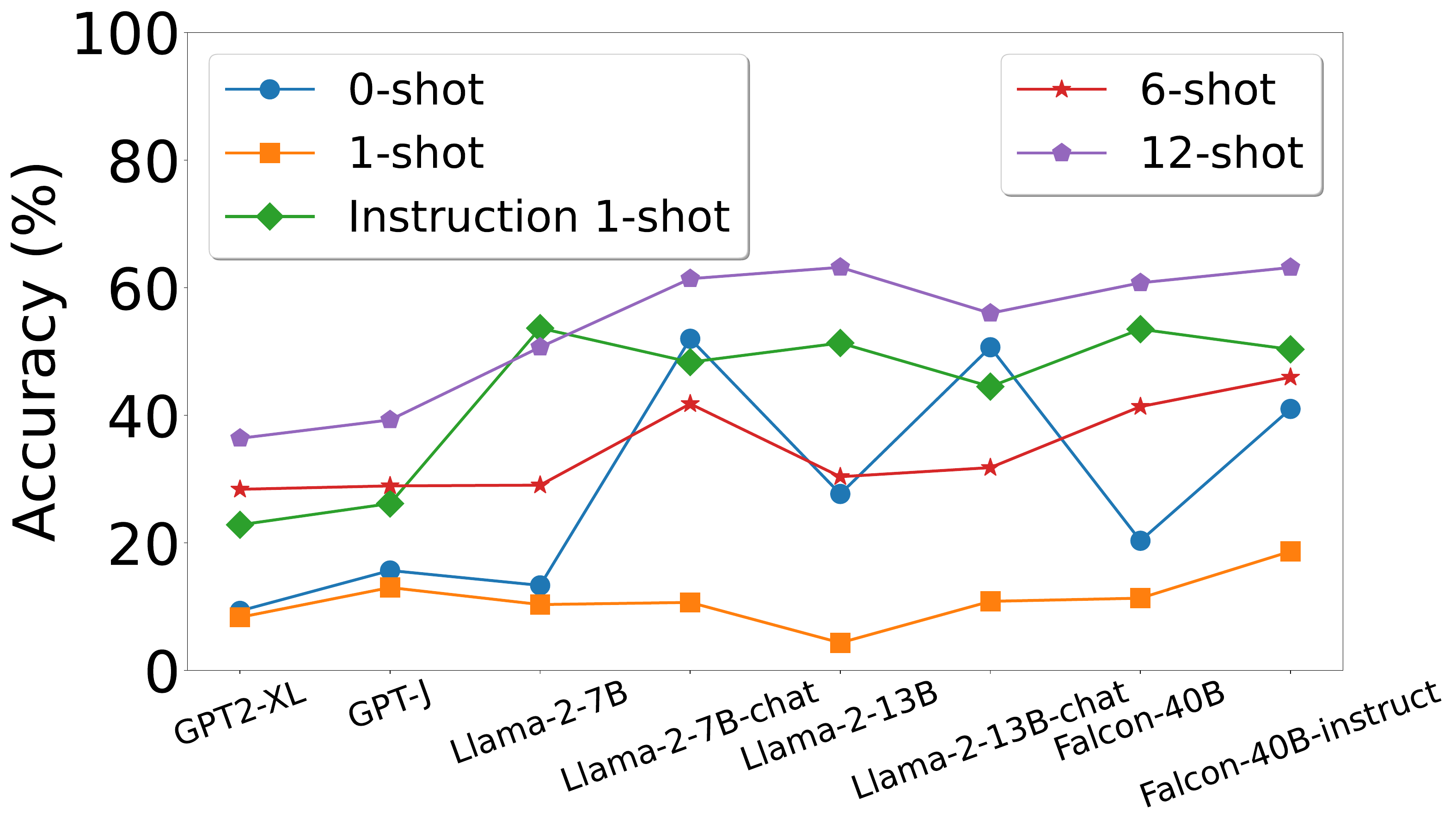}}
     \vspace*{-2em}
     \caption{\small The average accuracy of (1,6,12)-shot ICL w/o instructions and (0,1)-shot ICL with instructions.}
     \label{third_quadrant}
   \end{minipage}\hfill
   \begin{minipage}[t]{0.32\textwidth}
     \centering
     {\includegraphics[width=\linewidth]{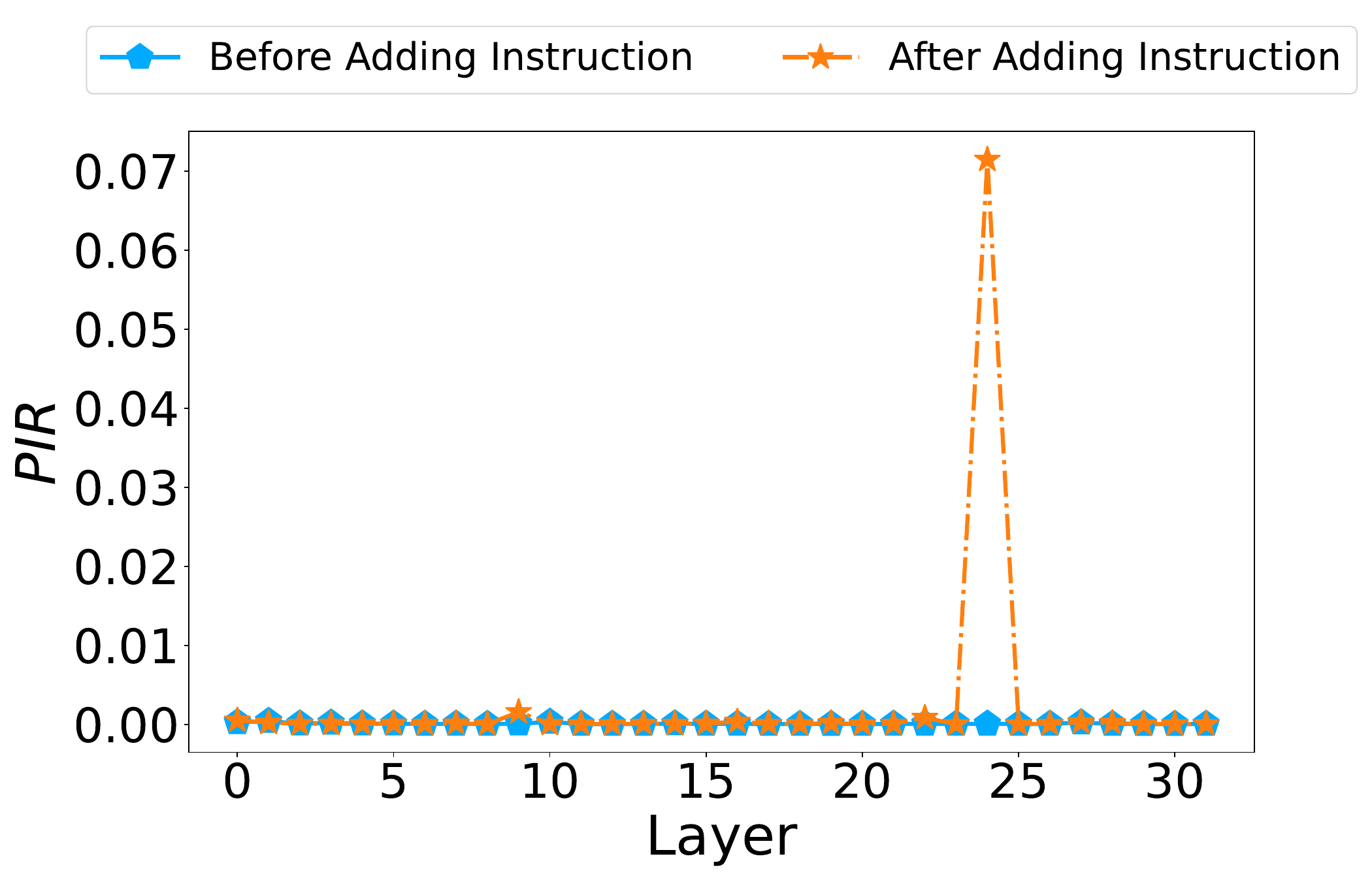}}
     \vspace*{-2em}
     \caption{\small The {\it \textbf{PIR}} of {\it "question"} at the label token {\it "Human"} on Llama-2-7B-chat, before and after adding instruction.}
     \label{instruction}
   \end{minipage}
\end{figure*}

\paragraph{Experimental Results.}
As illustrated in Figure \ref{fourth_quadrant}, the proportion of the model's predictions being the same as the incorrect label of \texttt{$\text{Similiar}({\text{T}}$}) is high. 
Notably, as the model size increases, this proportion reaches an exceptionally high level, indicating that the model almost entirely relies on the label of \texttt{$\text{Similiar}({\text{T}}$}) for its predictions.
We posit that this phenomenon arises because, as the model size increases, its ability to discern the similarity of examples improves, thereby directing more attention to similar examples.

\mybox{{\bf Conclusion}}{gray!40}{gray!10}{In the fourth quadrant, models directly replicate the labels of similar examples. Therefore, the performance of ICL depends heavily on whether the labels of similar examples match the ground truth labels of test samples. Additionally, larger models are better at recognizing similar examples, which increases their tendency to copy the labels from these examples.}

\section{Effects of Label Correctness and Demonstration Shot Number}
\label{Effects of Label Correctness and Shot Number}
As previous research presents two conflicting views about the effects of label correctness and shot number to ICL, in this section, we provide a brief summary about how our proposed coordinate system can explain this conflict in a more principled way.

\paragraph{Label Correctness.}
As long as similar examples are present in the demonstrations, the correctness of their labels consistently plays a crucial role in determining ICL performance, as discussed in Section \ref{First Quadrant} and Section \ref{Fourth Quadrant}. 
\textcolor{black}{For dissimilar examples, when the ICL scenario is positioned above the x-axis, label correctness can impact the models' confidence in task recognition.
Conversely, when positioned below the x-axis, label correctness does not decisively influence the models' predictions. 
In the third quadrant, models predominantly predict the label of the first example, whereas in the fourth quadrant, models are more likely to replicate the labels of similar examples.}

\paragraph{Demonstration Shot Number.}
Above the x-axis, increasing the shot number does not significantly affect ICL performance, as the models primarily rely on their pre-trained knowledge to make predictions once they recognize the task. Additional shots merely reinforce the models' confidence in correctly identifying the task. However, below the x-axis, increasing the shot number significantly impacts ICL performance. The more shots there are, the higher the likelihood that the models will find more similar examples to refer to. This approach can transition ICL from the third quadrant to the fourth quadrant and enhance the likelihood of including more similar examples for reference within the fourth quadrant.

\section{How to Make ICL Work Effectively?}
\label{How to Make ICL Work Effectively}
From our proposed ICL coordinate system, it is evident that the effectiveness of ICL improves as the values of the x and y coordinates increase, moving towards the upper right quadrant. 
Conversely, as the values of the x and y coordinates decrease, moving towards the lower left quadrant, the effectiveness of ICL diminishes. 
These observations provide valuable insights into the enhancement of ICL performance. 
Specifically, improvements can be made by:
\textbf{(1) strengthening the confidence in task recognition}, and 
\textbf{(2) providing examples with higher similarity in the demonstrations.}
We select the most challenging third quadrant to demonstrate how ICL can be made effective through these two directions.

\paragraph{Including a task description instruction before ICL examples can facilitate task recognition.} 
As illustrated in Figure \ref{instruction}, we conduct a one-shot ICL case study on the TREC dataset using Llama-2-7B-chat (for specific instructions and details on the one-shot ICL, refer to Appendix \ref{Specific Instructions and Details on One-Shot ICL}).
Without the task description instruction, the \textbf{\textit{PIR}} of the task-representative token {\it {"question"}} at the label is nearly zero.
However, after adding the task description instruction, the \textbf{\textit{PIR}} increases to 0.083 (ranking 12th in the vocabulary distribution), suggesting that the model recognizes the task to a certain extent.
This demonstrates that instructions can promote task recognition. 
This finding provides methodological support for the implementation of the first direction.
As shown in Figure \ref{third_quadrant}, the effectiveness of instruction one-shot ICL is significantly better than that of one-shot ICL without instructions.

\paragraph{Both retrieval and long-context ICL can provide examples with higher similarity.}
For the second direction, a commonly utilized method in recent research is to retrieve a highly similar subset of examples to serve as demonstrations for each test set example, which has been shown to be effective \citep{liu2021makes,rubin-etal-2022-learning}.
Additionally, as the context lengths of LLMs continue to increase, another method to achieve this goal is to continuously increase the number of input-label pairs in the demonstrations. 
The more input-label pairs included, the higher the likelihood that the model will find more similar examples to refer to during ICL.
As shown in Figure \ref{third_quadrant}, the performance of one-shot ICL is consistently worse than zero-shot for all models. 
However, each time we increase the number of shots, the performance of ICL improves significantly.
This aligns with the conclusions of \citet{bertsch2024incontext}, who find that long-context ICL can be surprisingly effective, with most of the improvement stemming from attending to similar examples rather than task recognition.

\begin{figure*}[t]
    \centering
    \begin{subfigure}[b]{0.3\textwidth}
        \includegraphics[width=\textwidth]{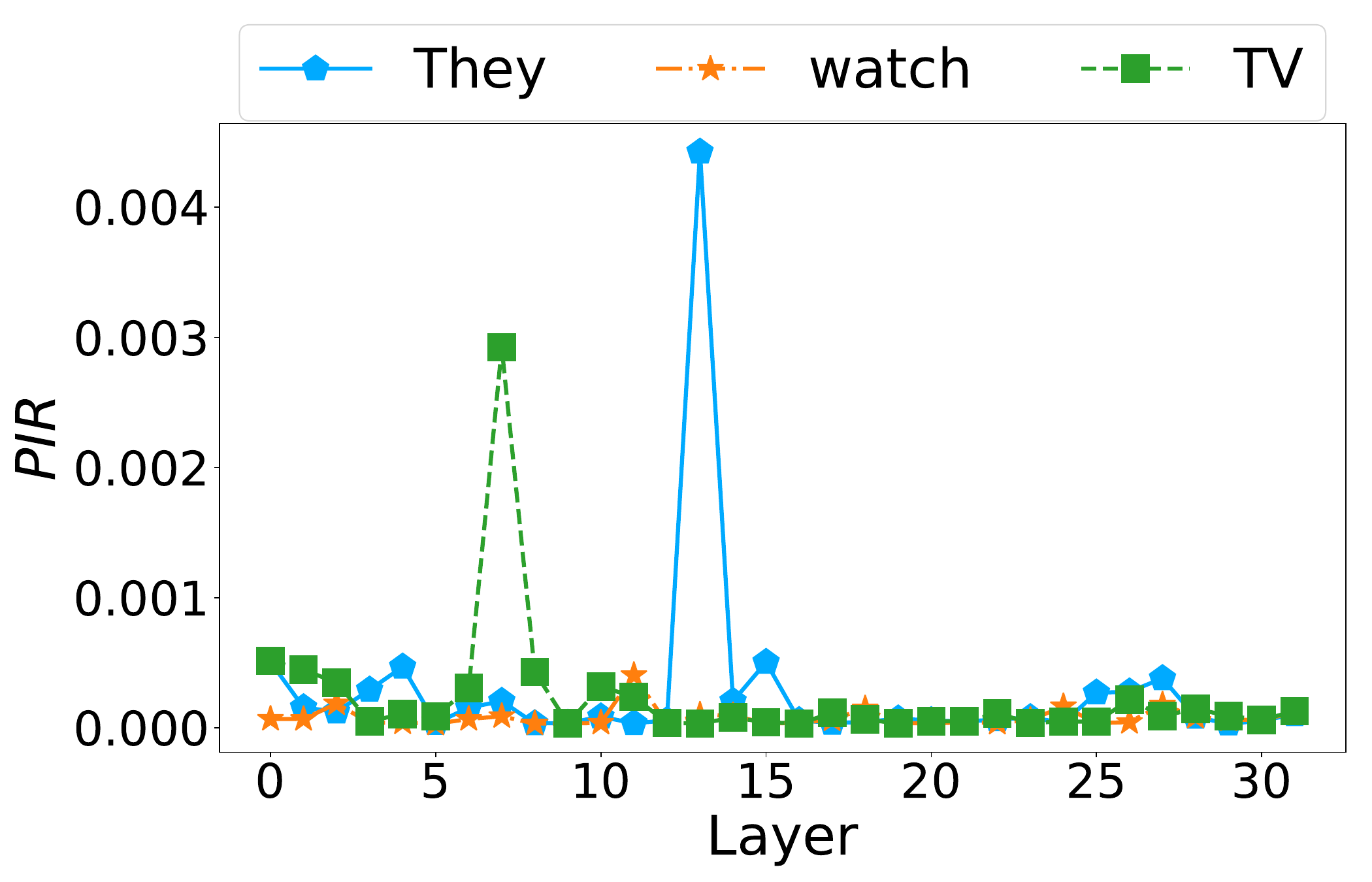}
         \vspace*{-1em}
        \caption{\small {\it \textbf{PIR}} of {\it "subject"} on each token of the final label when predicting {\it "He"}.}
        \label{fig:subject}
    \end{subfigure}
    \hfill
    \begin{subfigure}[b]{0.3\textwidth}
        \includegraphics[width=\textwidth]{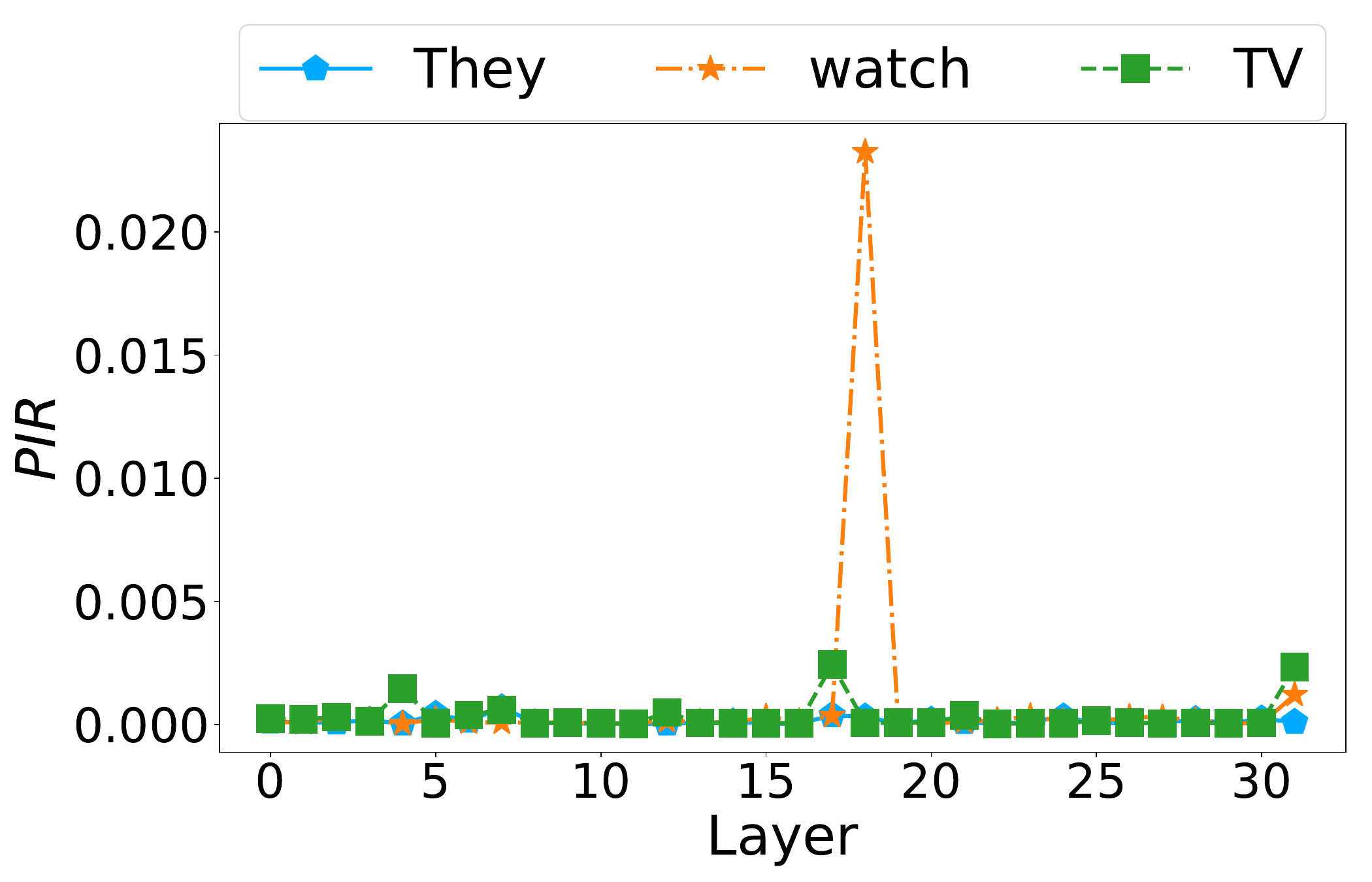}
         \vspace*{-1em}
        \caption{\small {\it \textbf{PIR}} of {\it "verb"} on each token of the final label when predicting {\it "reads"}.}
        \label{fig:verb}
    \end{subfigure}
    \hfill
    \begin{subfigure}[b]{0.3\textwidth}
        \includegraphics[width=\textwidth]{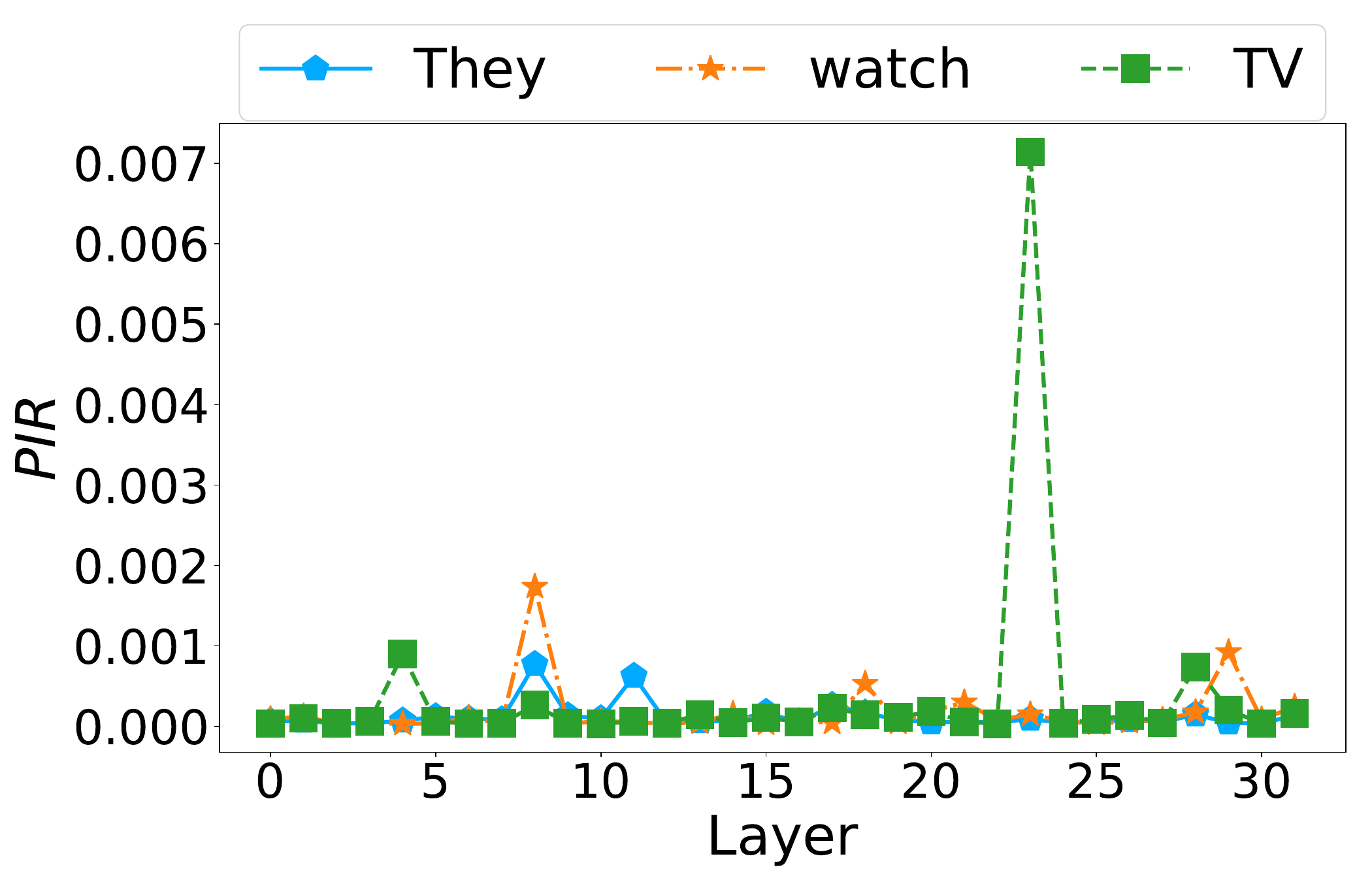}
         \vspace*{-1em}
        \caption{\small {\it \textbf{PIR}} of {\it "object"} on each token of the final label when predicting {\it "book"}.}
        \label{fig:object}
    \end{subfigure}
     \vspace*{-0.8em}
    \caption{\small For the in-context translation task with a strict subject-verb-object structure:
    \begin{CJK}{UTF8}{gbsn}
    "Sentence: 她喝水。 Label: She drinks water. Sentence: 我们吃米饭。 Label: We eat rice. Sentence: 他们看电视。 Label: They watch TV. Sentence: 他读书。 Label:"
    \end{CJK}, 
    the {\it \textbf{PIR}} of {\it "subject,"} {\it "verb,"} {\it "object"} at each token of the final label sentence when respectively predicting \textit{"He,"} \textit{"reads,"} \textit{"book,"} on Llama-2-7B.}
    \label{translation}
\end{figure*}

\section{Extension to Generation Tasks}

Given the recent success of ICL in generation tasks \citep{agrawal2022incontext, sia2023incontext, garcia2023unreasonable}, we aim for our two-dimensional coordinate system to enhance the understanding of ICL behavior not only in classification tasks but also in generation tasks.
This is non-trivial, as almost no prior work has conducted an in-depth analysis of in-context generation tasks.

To extend our coordinate system to generation tasks, we face two main challenges:
(1) In classification tasks, predicting a single label token is sufficient, whereas generation tasks require predicting an entire sentence.
(2) We determine whether ICL recognizes the task by examining if the hidden states of label tokens at internal layers possess task semantics. 
However, for generation tasks, the label for each example in the demonstrations is not a single word but a complete sentence.
To address these challenges, we formulate a hypothesis, $H$:
\begin{hypothesis}
    We treat a generation task as multiple smaller sub-classification tasks focused on predicting each token. 
    In each sub-classification task, every label sentence contains a token that serves as the label token.
    However, the position of the label token in each label sentence changes when predicting different tokens.
\end{hypothesis}

According to this hypothesis, each ICL sub-classification task in generation tasks is equivalent to the standard ICL in classification tasks we previously studied, as both involve predicting a single token, with the label token still represented by a single token.
However, verifying our hypothesis $H$ for most generative tasks is challenging. 
In classification tasks, the task of predicting each token (i.e., a label token) is very clear and specific. 
For instance, in SST-2, it is a \textit{{"sentiment"}} task. 
In contrast, for generative tasks, predicting a token may involve mimicking the abstract style of a segment of an input-label pair, or in more extreme cases, predicting just a complete article. 
Explicitly describing all the tasks that are involved in predicting these tokens is very difficult.

For this reason, we consider an in-context translation task, with all examples in the demonstrations adhering to a strict {\it subject-verb-object} structure, as illustrated in Figure \ref{translation}. If the ICL output also conforms to this strict subject-verb-object structure, it indicates that the tokens \textit{"He"}, \textit{"reads"}, and \textit{"book"} are generated by performing the explicit sub-tasks of {\it "subject"}, \textit{"verb"}, and {\it "object"}, respectively.
We analyze the label sentence of the last example in the demonstration. 
Figure \ref{translation} illustrates that for the subject token \textit{"He"}, the label token is \textit{"They"}, with the other tokens rarely generating {\it "subject"} semantics.
For the verb token \textit{"reads"}, the label token is \textit{"watch"}, with the other tokens seldom generating {\it "verb"} semantics. 
Similarly, for the object token \textit{"book"}, the label token is \textit{"TV"}, and the other tokens rarely generate {\it "object"} semantics.
This phenomenon of label token sliding within the label sentence provides evidence supporting the plausibility of our hypothesis \( H \). 

Therefore, by decomposing the entire generative task into multiple sub-classification tasks, our coordinate system can help understand the working mechanisms of in-context generative tasks.

\section{Related Work}

Current research on ICL mechanisms mainly falls into the following categories:

\paragraph{Theoretical Framework.}
Recently, numerous studies have employed theoretical frameworks to enhance the understanding of ICL.
\citet{xie2022an} describe ICL as implicit Bayesian inference. \citet{garg2023transformers} demonstrate that transformers can learn linear functions through ICL. 
Additionally, several studies conceptualize ICL as gradient descent on an implicit internal model \citep{akyurek2022learning,von2023transformers,dai2022can}.
\citet{wei2023larger} and \citet{pan2023context} disentangle ICL into task recognition and task learning.

\paragraph{Empiricism.} Various factors affecting ICL have been studied, such as the order of examples \citep{lu2022fantastically}, the choice of label words \citep{min2022rethinking,yoo2022groundtruth}, and the selection of demonstrations \citep{liu2021makes,rubin-etal-2022-learning}. Effective demonstration strategies \citep{ye2023compositional,li2023unified} can notably boost ICL performance.

\paragraph{Logit Lens.}
Recently, some studies use the logit lens technique \citep{nostalgebraist2020logit,geva2021transformer} to project complex feature representations into the vocabulary space to study the mechanisms of ICL. 
\citet{merullo2023mechanism} discover three distinct stages of processing by decoding the next token prediction at each layer. 
\citet{todd2024function} find that a small number of attention heads transport a compact representation of the demonstrated task. 
\citet{yu2024large} provide insights into the mechanisms of ICL in the context of task learning.

\section{Conclusion}
In this paper, we map two variables {\it whether LLMs can recognize the task} and {\it the presence of similar examples in the demonstrations} onto the y-axis and x-axis of a 2D coordinate system, to visualize ICL scenarios. First, for classification tasks, we conduct a systematic study of the proposed coordinate system and describe in detail the working mechanisms of ICL in each quadrant. Then, we extend our analyses beyond classification tasks through a thorough case study on a machine translation task. Our proposed coordinate system offers a universal framework to better understand ICL.


\section*{Limitations}

Our extensive studies, despite offering a principled and universal approach to understanding the working mechanism of ICL, have several limitations. First, our research primarily focused on conventional ICL paradigms, leaving other paradigms such as chain of thought prompting (CoT) \citep{wei2023chainofthought} unexplored. Second, for generative tasks, we conducted a case study solely on in-context machine translation tasks adhering to a strict subject-verb-object structure. Third, due to hardware constraints, our investigation was primarily limited to models with up to 40 billion parameters. Further research replicating our study could use larger models with our 2D coordinate system to uncover more interesting findings.

\section*{Acknowledgments}
This research / project is supported by the National Research Foundation, Singapore under its Industry Alignment Fund – Pre-positioning (IAF-PP) Funding Initiative. Any opinions, findings and conclusions or recommendations expressed in this material are those of the author(s) and do not reflect the views of National Research Foundation, Singapore. We thank Xingluan (AI Cloud computing service), EIT and IDT High Performance Computing Center for providing computational resources for this project.
This work corresponds to Fanghua Ye, Jinlan Fu, and Xiaoyu Shen.

\bibliography{custom}

\clearpage 

\appendix

\section{Lexical Similarity or Semantical Simlarity}
\label{Similarity}
In this section, we explore whether models give significantly more attention to semantically similar examples (which delve into deeper meanings) or to lexically similar examples (which may have opposite semantics but share superficial similarities). We also consider randomly selected dataset examples as a baseline. 
For this purpose, in the context of task learning (a setup first introduced by \citet{pan2023context}, which replaces all labels with semantically irrelevant words to prevent ICL from recognizing the task), we include three elements in the demonstrations, each representing a different type of similarity to the test sample, and observe which element's label the model predicts. The definitions of these three elements are as follows:
\begin{itemize}

\item {\textbf{$\text{Example}_{\text{lexical}}$}:} An example that largely overlaps lexically with the test sample, yet differs semantically and carries a distinct label.

\item {\textbf{$\text{Example}_{\text{semantic}}$}:} An example that is semantically similar to the test sample, but has minimal lexical overlap and carries the same label. Specifically, we achieve this by paraphrasing the test sample.

\item {\textbf{$\text{Example}_{\text{baseline}}$}:} A randomly selected example from the dataset, with minimal lexical and semantic similarity to the test sample.
\end{itemize}

Notably, we find that when each element appears only once in the demonstrations, as in a 3-shot ICL scenario, models tend to generalize the pattern of label changes. For instance, if the three labels are sequentially "a," "b," and "c," the models are likely to predict "d." To mitigate this phenomenon, we replicate each element multiple times (in our experiments, three times). Empirical experiments show that this approach can ensure the models' predictions stay within the intended label space.

\paragraph{\textbf{Models}.} We employ various models from the GPT series, including GPT2-Medium (355M) and GPT2-XL (1.61B) \citep{radford2019language}, as well as GPT-J (6B) \citep{mesh-transformer-jax}. To investigate the impact of instruction-tuning on similarity preferences, we also employ Llama-2-7B \citep{touvron2023llama}, Llama-3-8B \citep{llama3modelcard}, and Mistral-7B-v0.1 \citep{jiang2023mistral}, along with their instruction-tuned versions. All checkpoints of these models are sourced from the transformers library \citep{wolf2019huggingface}.

\paragraph{\textbf{Datasets}.}  We adopt the Stanford Sentiment Treebank Binary (SST-2) \citep{socher-etal-2013-recursive} for sentiment analysis, Text REtrieval Conference Question Classification (TREC) (\citealp{li2002learning}; \citealp{hovy2001toward}) for question type classification, EmoContext (emo) \citep{chatterjee-etal-2019-semeval} for emotion classification, and \texttt{hate\_speech18} \citep{gibert2018hate} for hate speech detection.
For each dataset, we leverage GPT-4 \citep{openai2024gpt4} to generate 20 triplets, formatted as \texttt{($\text{Test Sample}$, $\text{Example}_{\text{semantic}}$, $\text{Example}_{\text{lexical}}$)}, that are tailored to the style of the dataset. 
After generation, these triplets undergo a manual selection process to ensure quality.
In addition, for each triplet, we randomly select an example from the original dataset to serve as \texttt{$\text{Example}_{\text{baseline}}$}, thereby forming a complete demonstration.
For detailed information on the triplets generated for each dataset, please refer to Appendix \ref{Triplets}.

\begin{figure*}[!th]
  \includegraphics[width=0.95\textwidth]{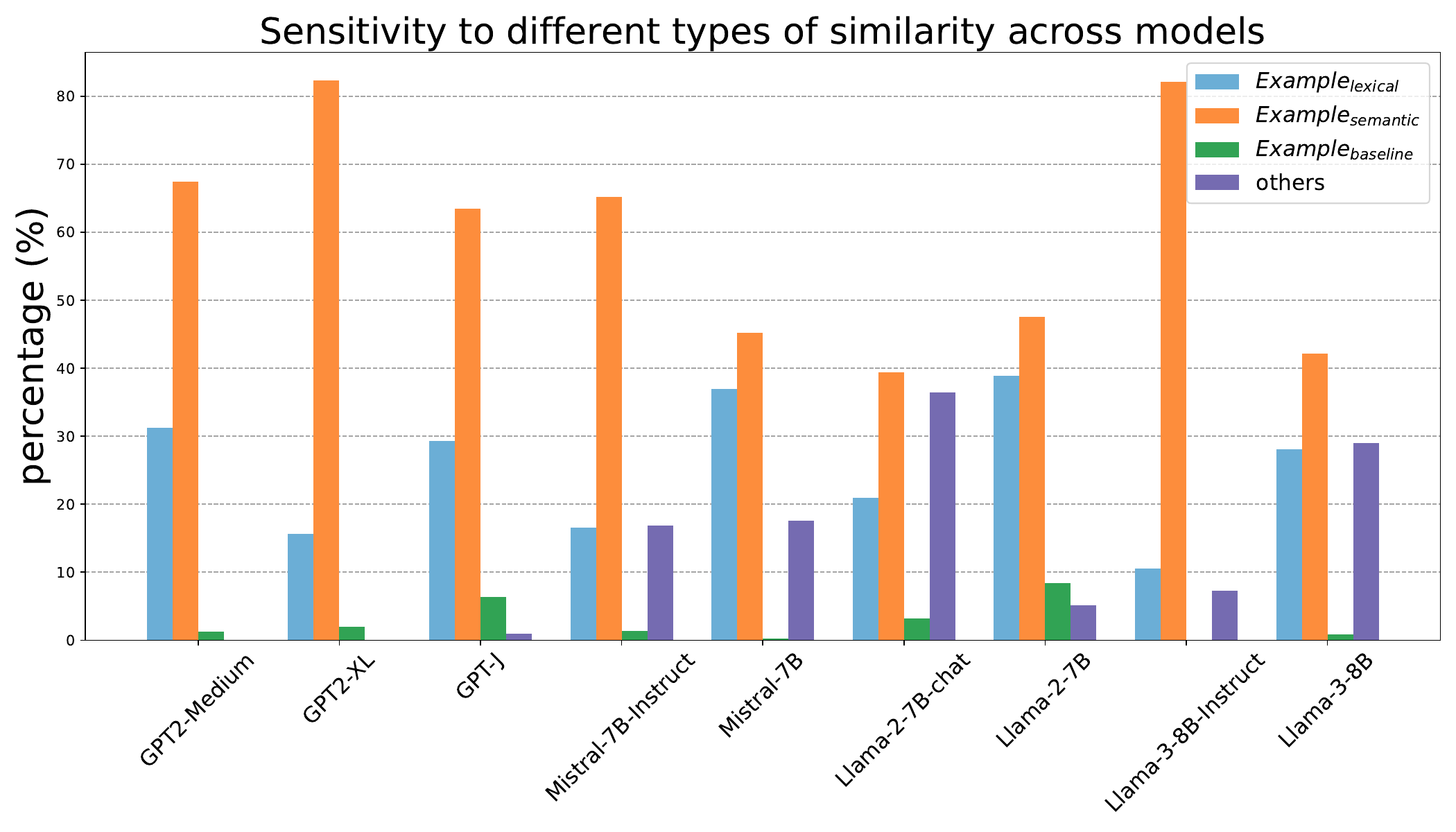}
  \caption{The proportion of each element's corresponding label in all predictions. Although we adopt the method of repeating each example in the demonstrations to mitigate the models' tendency to summarize patterns of label changes when performing ICL, for some models, it is still inevitable that tokens outside the label space are predicted. We use "others" to represent all tokens outside the label space.
}
  \label{bar_chart_similarity}
\end{figure*}

\paragraph{\textbf{Implementation Details}.} In our experiments, the prompts consist solely of demonstrations without incorporating instructions. We employ neutral delimiters, specifically "Sentence:" and "Label:", to clearly separate the components of the demonstrations.
This approach ensures that the models do not receive any task-specific information that could be inferred from the delimiters.
Although previous works in the task learning setting typically employ semantically irrelevant words as labels, such as "foo" and "bar," we choose to label the elements with the initial letters of "lexical," "semantic," and "baseline" for ease of distinction. Consequently, the labels used for \texttt{$\text{Example}_{\text{lexical}}$}, \texttt{$\text{Example}_{\text{semantic}}$},  and \texttt{$\text{Example}_{\text{baseline}}$} are "l," "s," and "b," respectively.
For each ICL prediction, we document the token that ranked the highest in the model's output distribution. Subsequently, across all predictions, we calculate the proportions of the tokens "l," "s," and "b" respectively.
The experimental results are averaged over five random seeds and all datasets.

\paragraph{\textbf{Results and Analysis}.}
The experimental results, as shown in Figure \ref{bar_chart_similarity}, reveal that:
(1) Models pay significantly more attention to examples with lexical similarity or semantic similarity compared to randomly selected examples with minimal lexical similarity and semantic similarity.
(2) Across all models, attention to semantic similarity is higher than to lexical similarity.
(3) As the model size increases, models are more likely to predict tokens outside the label space, indicating that larger models place greater weight on summarizing patterns of label changes in the context of task learning.
(4) Instruction-tuned models tend to favor examples with semantic similarity, possibly due to the emphasis on semantic understanding during the instruction-tuning training process.

\section{GPT-4 Generated Triplets}
\label{Triplets}
Refer to Tables \ref{sst2-part1} and \ref{sst2-part2} for the 20 triplets generated by GPT-4 on the SST-2 dataset. Refer to Tables \ref{emo-part1} and \ref{emo-part2} for the 20 triplets generated by GPT-4 on the emo dataset. Refer to Tables \ref{trec-part1} and \ref{trec-part2} for the 20 triplets generated by GPT-4 on the TREC dataset. Refer to Tables \ref{hate-part1} and \ref{hate-part2} for the 20 triplets generated by GPT-4 on the \texttt{hate\_speech18} dataset. 

\begin{table*}[th!]
\centering
\captionsetup{justification=centering}
\begin{tabular}{|p{5cm}|p{5cm}|p{5cm}|}
\toprule
\textbf{Test sample} & \textbf{$\text{Example}_{\text{semantic}}$} & \textbf{$\text{Example}_{\text{lexical}}$} \\
\midrule
The soundtrack enriches the entire movie. \textbf{Label: Positive} & The music significantly enhances the film's appeal. \textbf{Label: Positive} & The soundtrack diminishes the entire movie. \textbf{Label: Negative} \\
\addlinespace
The actor's performance is truly mesmerizing. \textbf{Label: Positive} & The actor's portrayal captivates the audience completely. \textbf{Label: Positive} & The actor's performance is truly forgettable. \textbf{Label: Negative} \\
\addlinespace
The plot twists were unexpected and thrilling. \textbf{Label: Positive} & The story developments were surprising and exciting. \textbf{Label: Positive} & The plot twists were predictable and dull. \textbf{Label: Negative} \\
\addlinespace
The cinematography is breathtaking and innovative. \textbf{Label: Positive} & The visual direction offers stunning and groundbreaking visuals. \textbf{Label: Positive} & The cinematography is uninspiring and outdated. \textbf{Label: Negative} \\
\addlinespace
The dialogue was witty and delightful. \textbf{Label: Positive} & The conversation was sharp and enjoyable. \textbf{Label: Positive} & The dialogue was humorless and disappointing. \textbf{Label: Negative} \\
\addlinespace
The direction is masterful and precise. \textbf{Label: Positive} & The film's guidance shows exceptional skill and accuracy. \textbf{Label: Positive} & The direction is clumsy and imprecise. \textbf{Label: Negative} \\
\addlinespace
The special effects are spectacular and memorable. \textbf{Label: Positive} & The visual effects stand out as extraordinary and unforgettable. \textbf{Label: Positive} & The special effects are unimpressive and forgettable. \textbf{Label: Negative} \\
\addlinespace
The pacing keeps you engaged from start to finish. \textbf{Label: Positive} & The rhythm maintains your attention throughout the entire movie. \textbf{Label: Positive} & The pacing loses your interest from start to finish. \textbf{Label: Negative} \\
\addlinespace
The characters are richly developed and relatable. \textbf{Label: Positive} & The portrayal of characters is deeply crafted and connects well with the audience. \textbf{Label: Positive} & The characters are poorly developed and unrelatable. \textbf{Label: Negative} \\
\addlinespace
The film's creativity is both refreshing and inspiring. \textbf{Label: Positive} & The movie's originality offers a new and motivational perspective. \textbf{Label: Positive} & The film's creativity is both stale and uninspiring. \textbf{Label: Negative} \\
\bottomrule
\end{tabular}
\caption{For the SST-2 dataset, 20 triplets generated by GPT-4 (Part 1).}
\label{sst2-part1}
\end{table*}


\begin{table*}[th!]
\centering
\captionsetup{justification=centering}
\begin{tabular}{|p{5cm}|p{5cm}|p{5cm}|}
\toprule
\textbf{Test sample} & \textbf{$\text{Example}_{\text{semantic}}$} & \textbf{$\text{Example}_{\text{lexical}}$} \\
\midrule
I despise this show for its lack of originality. \textbf{Label: Negative} & The series annoys me with its derivative content. \textbf{Label: Negative} & I adore this show despite its lack of originality. \textbf{Label: Positive} \\
\addlinespace
The ending was predictable and boring. \textbf{Label: Negative} & The conclusion was foreseeable and tedious. \textbf{Label: Negative} & The ending was unpredictable and exciting. \textbf{Label: Positive} \\
\addlinespace
Their service was slow and frustrating. \textbf{Label: Negative} & The customer service was sluggish and irritating. \textbf{Label: Negative} & Their service was quick and satisfying. \textbf{Label: Positive} \\
\addlinespace
It's utterly pointless and dull. \textbf{Label: Negative} & Completely meaningless and uninteresting. \textbf{Label: Negative} & It's utterly purposeful and engaging. \textbf{Label: Positive} \\
\addlinespace
The plot twists were contrived and unconvincing. \textbf{Label: Negative} & The storyline turns felt forced and unbelievable. \textbf{Label: Negative} & The plot twists were natural and convincing. \textbf{Label: Positive} \\
\addlinespace
The movie was generally uninspiring. \textbf{Label: Negative} & The film rarely evoked any excitement. \textbf{Label: Negative} & The movie was generally inspiring. \textbf{Label: Positive} \\
\addlinespace
The soundtrack is hardly noticeable. \textbf{Label: Negative} & You barely hear the music throughout. \textbf{Label: Negative} & The soundtrack is highly noticeable. \textbf{Label: Positive} \\
\addlinespace
The pacing is slow and tedious. \textbf{Label: Negative} & The tempo drags and feels monotonous. \textbf{Label: Negative} & The pacing is quick and engaging. \textbf{Label: Positive} \\
\addlinespace
The narrative lacks depth and coherence. \textbf{Label: Negative} & The story misses complexity and clarity. \textbf{Label: Negative} & The narrative has depth and coherence. \textbf{Label: Positive} \\
\addlinespace
The performance was overly dramatic and false. \textbf{Label: Negative} & The acting was excessively theatrical and inauthentic. \textbf{Label: Negative} & The performance was subtly dramatic and genuine. \textbf{Label: Positive} \\
\bottomrule
\end{tabular}
\caption{For the SST-2 dataset, 20 triplets generated by GPT-4 (Part 2).}
\label{sst2-part2}
\end{table*}

\begin{table*}[th!]
\centering
\captionsetup{justification=centering}
\begin{tabular}{|p{5cm}|p{5cm}|p{5cm}|}
\toprule
\textbf{Test sample} & \textbf{$\text{Example}_{\text{semantic}}$} & \textbf{$\text{Example}_{\text{lexical}}$} \\
\midrule
It seems like a regular day at the office. \textbf{Label: Others} & Just another normal workday. \textbf{Label: Others} & Today, the office feels unsettlingly quiet. \textbf{Label: Sad} \\
\addlinespace
I need to go grocery shopping later. \textbf{Label: Others} & Later today, I have some grocery shopping to do. \textbf{Label: Others} & I'm frustrated about having to go grocery shopping later. \textbf{Label: Angry} \\
\addlinespace
I'm so happy we're going on a vacation! \textbf{Label: Happy} & I'm thrilled about our upcoming vacation! \textbf{Label: Happy} & I'm stressed about all the packing needed for our vacation. \textbf{Label: Angry} \\
\addlinespace
That birthday party was a blast! \textbf{Label: Happy} & I truly enjoyed the fun at that birthday party! \textbf{Label: Happy} & That birthday party was too loud and overwhelming for me. \textbf{Label: Sad} \\
\addlinespace
Losing my pet has left me heartbroken. \textbf{Label: Sad} & I am deeply saddened by the loss of my pet. \textbf{Label: Sad} & Dealing with my pet's loss has made me irritable and upset. \textbf{Label: Angry} \\
\addlinespace
It's so gloomy outside today, it makes me feel down. \textbf{Label: Sad} & The dreary weather today really dampens my spirits. \textbf{Label: Sad} & The gloomy weather outside is irritating. \textbf{Label: Angry} \\
\addlinespace
I can't believe how unfair that decision was! \textbf{Label: Angry} & I'm really upset about that unjust decision! \textbf{Label: Angry} & That decision was so disappointing and unfair. \textbf{Label: Sad} \\
\addlinespace
This constant noise is driving me crazy! \textbf{Label: Angry} & I'm getting furious over the incessant noise! \textbf{Label: Angry} & This constant noise is really getting on my nerves. \textbf{Label: Others} \\
\addlinespace
I can't believe I got promoted at work! \textbf{Label: Happy} & I am so excited about my promotion at work! \textbf{Label: Happy} & I can't believe how stressed I am at work. \textbf{Label: Sad} \\
\addlinespace
Missing the bus has ruined my day. \textbf{Label: Sad} & Missing the bus completely ruined my entire day. \textbf{Label: Sad} & Missing the bus has made me furious. \textbf{Label: Angry} \\
\bottomrule
\end{tabular}
\caption{For the emo dataset, 20 triplets generated by GPT-4 (Part 1).}
\label{emo-part1}
\end{table*}


\begin{table*}[th!]
\centering
\captionsetup{justification=centering}
\begin{tabular}{|p{5cm}|p{5cm}|p{5cm}|}
\toprule
\textbf{Test sample} & \textbf{$\text{Example}_{\text{semantic}}$} & \textbf{$\text{Example}_{\text{lexical}}$} \\
\midrule
My friend surprised me with a gift. \textbf{Label: Happy} & I was delighted when my friend gave me a gift. \textbf{Label: Happy} & My friend surprised me with a rude comment. \textbf{Label: Angry} \\
\addlinespace
The news of the accident left me in tears. \textbf{Label: Sad} & I cried when I heard about the accident. \textbf{Label: Sad} & The news of the accident left me feeling numb. \textbf{Label: Others} \\
\addlinespace
I'm so angry that my computer crashed again! \textbf{Label: Angry} & It's infuriating that my computer crashed again! \textbf{Label: Angry} & I'm so sad that my computer crashed again. \textbf{Label: Sad} \\
\addlinespace
It's a beautiful day outside, I feel great. \textbf{Label: Happy} & The nice weather outside makes me feel wonderful. \textbf{Label: Happy} & It's a beautiful day outside, but I feel anxious. \textbf{Label: Others} \\
\addlinespace
The way they treated me was so disrespectful. \textbf{Label: Angry} & Their treatment of me was incredibly disrespectful. \textbf{Label: Angry} & The way they treated me was so kind. \textbf{Label: Happy} \\
\addlinespace
I'm feeling down because I didn't get the job. \textbf{Label: Sad} & I'm feeling really sad because I didn't get hired. \textbf{Label: Sad} & I'm feeling great because I got the job. \textbf{Label: Happy} \\
\addlinespace
This project has been so rewarding. \textbf{Label: Happy} & Working on this project has been incredibly fulfilling. \textbf{Label: Happy} & This project has been so frustrating. \textbf{Label: Angry} \\
\addlinespace
I can't stand the traffic jam every morning. \textbf{Label: Angry} & The daily traffic jam every morning drives me nuts. \textbf{Label: Angry} & I love the peaceful mornings without traffic jams. \textbf{Label: Happy} \\
\addlinespace
I'm ecstatic about the new opportunities ahead! \textbf{Label: Happy} & I'm thrilled about the upcoming new opportunities! \textbf{Label: Happy} & I'm anxious about the new challenges ahead. \textbf{Label: Sad} \\
\addlinespace
I can't believe they forgot my birthday. \textbf{Label: Sad} & It's disappointing that they forgot my birthday. \textbf{Label: Sad} & I can't believe they remembered my birthday. \textbf{Label: Happy} \\
\bottomrule
\end{tabular}
\caption{For the emo dataset, 20 triplets generated by GPT-4 (Part 2).}
\label{emo-part2}
\end{table*}

\begin{table*}[th!]
\centering
\begin{tabular}{|p{5cm}|p{5cm}|p{5cm}|}
\toprule
\textbf{Test sample} & \textbf{$\text{Example}_{\text{semantic}}$} & \textbf{$\text{Example}_{\text{lexical}}$} \\
\midrule
What does 'CPU' stand for? \textbf{Label: ABBR} & What is the meaning of 'CPU'? \textbf{Label: ABBR} & Where is the 'CPU' located? \textbf{Label: LOC} \\
\addlinespace
What is the capital of France? \textbf{Label: LOC} & Which city is the capital of France? \textbf{Label: LOC} & What is the population of France? \textbf{Label: NUM} \\
\addlinespace
Who discovered electricity? \textbf{Label: HUM} & Name the person who discovered electricity. \textbf{Label: HUM} & When was electricity discovered? \textbf{Label: NUM} \\
\addlinespace
What is the boiling point of water? \textbf{Label: NUM} & At what temperature does water boil? \textbf{Label: NUM} & Who discovered the boiling point of water? \textbf{Label: HUM} \\
\addlinespace
Who is the CEO of Tesla? \textbf{Label: HUM} & Identify the current CEO of Tesla. \textbf{Label: HUM} & Where is the headquarters of Tesla? \textbf{Label: LOC} \\
\addlinespace
What does 'HTTP' mean? \textbf{Label: ABBR} & Explain the term 'HTTP'. \textbf{Label: ABBR} & Who invented 'HTTP'? \textbf{Label: HUM} \\
\addlinespace
Name a famous painter from Spain. \textbf{Label: HUM} & Who is a renowned Spanish painter? \textbf{Label: HUM} & What is a famous painting from Spain? \textbf{Label: ENTY} \\
\addlinespace
Describe the process of photosynthesis. \textbf{Label: DESC} & Explain how photosynthesis works. \textbf{Label: DESC} & Who discovered photosynthesis? \textbf{Label: HUM} \\
\addlinespace
Where is the Great Wall of China located? \textbf{Label: LOC} & Locate the Great Wall of China. \textbf{Label: LOC} & When was the Great Wall of China built? \textbf{Label: NUM} \\
\addlinespace
How many continents are there? \textbf{Label: NUM} & How many continents exist? \textbf{Label: NUM} & What is the largest continent? \textbf{Label: ENTY} \\
\bottomrule
\end{tabular}
\caption{For the TREC dataset, 20 triplets generated by GPT-4 (Part 1), where ENTY stands for Entity, HUM stands for Human being, NUM stands for Numeric value, LOC stands for Location, ABBR stands for Abbreviation, and DESC stands for Description and abstract concept.
}
\label{trec-part1}
\end{table*}


\begin{table*}[th!]
\centering
\begin{tabular}{|p{5cm}|p{5cm}|p{5cm}|}
\toprule
\textbf{Test sample} & \textbf{$\text{Example}_{\text{semantic}}$} & \textbf{$\text{Example}_{\text{lexical}}$} \\
\midrule
What is the full form of 'UNICEF'? \textbf{Label: ABBR} & What does the abbreviation 'UNICEF' represent? \textbf{Label: ABBR} & Who founded 'UNICEF'? \textbf{Label: HUM} \\
\addlinespace
Who won the Nobel Peace Prize in 2020? \textbf{Label: HUM} & Who was awarded the Nobel Peace Prize in 2020? \textbf{Label: HUM} & What is the prize money for the Nobel Peace Prize? \textbf{Label: NUM} \\
\addlinespace
What is the meaning of 'quantum physics'? \textbf{Label: DESC} & Define 'quantum physics'. \textbf{Label: DESC} & Who coined the term 'quantum physics'? \textbf{Label: HUM} \\
\addlinespace
Name the author of 'Pride and Prejudice'. \textbf{Label: HUM} & Identify the writer of 'Pride and Prejudice'. \textbf{Label: HUM} & When was 'Pride and Prejudice' published? \textbf{Label: NUM} \\
\addlinespace
What is the largest ocean on Earth? \textbf{Label: ENTY} & Which ocean is the largest on Earth? \textbf{Label: ENTY} & Where is the largest ocean located? \textbf{Label: LOC} \\
\addlinespace
How tall is Mount Everest? \textbf{Label: NUM} & What is the height of Mount Everest? \textbf{Label: NUM} & Who measured the height of Mount Everest? \textbf{Label: HUM} \\
\addlinespace
What does 'DNA' stand for? \textbf{Label: ABBR} & Explain what 'DNA' stands for. \textbf{Label: ABBR} & Where is 'DNA' found? \textbf{Label: LOC} \\
\addlinespace
What type of animal is a tiger? \textbf{Label: ENTY} & What species does a tiger belong to? \textbf{Label: ENTY} & Where are tigers commonly found? \textbf{Label: LOC} \\
\addlinespace
Explain the theory of relativity. \textbf{Label: DESC} & Describe the concept of relativity. \textbf{Label: DESC} & Who developed the theory of relativity? \textbf{Label: HUM} \\
\addlinespace
Who was the first woman to fly solo across the Atlantic? \textbf{Label: HUM} & Who was the first female to fly solo across the Atlantic? \textbf{Label: HUM} & When did the first woman fly solo across the Atlantic? \textbf{Label: NUM} \\
\bottomrule
\end{tabular}
\caption{For the TREC dataset, 20 triplets generated by GPT-4 (Part 2), where ENTY stands for Entity, HUM stands for Human being, NUM stands for Numeric value, LOC stands for Location, ABBR stands for Abbreviation, and DESC stands for Description and abstract concept.}
\label{trec-part2}
\end{table*}

\begin{table*}[th!]
\centering
\captionsetup{justification=centering}
\begin{tabular}{|p{5cm}|p{5cm}|p{5cm}|}
\toprule
\textbf{Test sample} & \textbf{$\text{Example}_{\text{semantic}}$} & \textbf{$\text{Example}_{\text{lexical}}$} \\
\midrule
You are such an inspiration to everyone around you. \textbf{Label: noHate} & You inspire everyone around you. \textbf{Label: noHate} & You are such a burden to everyone around you. \textbf{Label: hate} \\
\addlinespace
I hate it when people like you try to ruin everything. \textbf{Label: hate} & I can't stand it when people like you mess everything up. \textbf{Label: hate} & I love it when people like you try to make things better. \textbf{Label: noHate} \\
\addlinespace
Your kindness and generosity know no bounds. \textbf{Label: noHate} & Your endless kindness and generosity are remarkable. \textbf{Label: noHate} & Your selfishness and greed know no bounds. \textbf{Label: hate} \\
\addlinespace
People like you make this world a terrible place. \textbf{Label: hate} & People like you make this world worse. \textbf{Label: hate} & People like you make this world a wonderful place. \textbf{Label: noHate} \\
\addlinespace
I'm so grateful for your support and friendship. \textbf{Label: noHate} & I'm deeply thankful for your support and friendship. \textbf{Label: noHate} & I'm so resentful of your support and friendship. \textbf{Label: hate} \\
\addlinespace
You are nothing but a waste of space. \textbf{Label: hate} & You are completely useless. \textbf{Label: hate} & You are everything but a waste of space. \textbf{Label: noHate} \\
\addlinespace
Your efforts are making a significant difference. \textbf{Label: noHate} & Your contributions are having a major impact. \textbf{Label: noHate} & Your efforts are making no difference at all. \textbf{Label: hate} \\
\addlinespace
You don't belong here, go back to where you came from. \textbf{Label: hate} & You should leave and never come back. \textbf{Label: hate} & You belong here, stay where you are. \textbf{Label: noHate} \\
\addlinespace
Thank you for always being there for me. \textbf{Label: noHate} & Thanks for always being supportive. \textbf{Label: noHate} & Thank you for never being there for me. \textbf{Label: hate} \\
\addlinespace
You are the reason why nothing ever gets better. \textbf{Label: hate} & You are why things never improve. \textbf{Label: hate} & You are the reason why everything gets better. \textbf{Label: noHate} \\
\bottomrule
\end{tabular}
\caption{For the \texttt{hate\_speech18} dataset, 20 triplets generated by GPT-4 (Part 1).
}
\label{hate-part1}
\end{table*}


\begin{table*}[th!]
\centering
\begin{tabular}{|p{5cm}|p{5cm}|p{5cm}|}
\toprule
\textbf{Test sample} & \textbf{$\text{Example}_{\text{semantic}}$} & \textbf{$\text{Example}_{\text{lexical}}$} \\
\midrule
Your dedication to helping others is commendable. \textbf{Label: noHate} & Your commitment to helping others is admirable. \textbf{Label: noHate} & Your dedication to harming others is commendable. \textbf{Label: hate} \\
\addlinespace
I can't stand people like you. \textbf{Label: hate} & People like you are unbearable. \textbf{Label: hate} & I can't stand people like you making a positive impact. \textbf{Label: noHate} \\
\addlinespace
You have a heart of gold and it shows in everything you do. \textbf{Label: noHate} & Your golden heart is evident in all your actions. \textbf{Label: noHate} & You have a heart of stone and it shows in everything you do. \textbf{Label: hate} \\
\addlinespace
Why don't you just disappear already? \textbf{Label: hate} & Why don't you just go away? \textbf{Label: hate} & Why don't you just stay and make things better? \textbf{Label: noHate} \\
\addlinespace
I'm amazed by your talent and hard work. \textbf{Label: noHate} & Your talent and hard work are truly impressive. \textbf{Label: noHate} & I'm amazed by your lack of talent and hard work. \textbf{Label: hate} \\
\addlinespace
You're nothing but a burden to everyone. \textbf{Label: hate} & You're a burden to everyone around you. \textbf{Label: hate} & You're nothing but a blessing to everyone. \textbf{Label: noHate} \\
\addlinespace
Your positive attitude is contagious. \textbf{Label: noHate} & Your positive outlook is infectious. \textbf{Label: noHate} & Your negative attitude is contagious. \textbf{Label: hate} \\
\addlinespace
You are a disgrace to your family. \textbf{Label: hate} & You are a shame to your family. \textbf{Label: hate} & You are an honor to your family. \textbf{Label: noHate} \\
\addlinespace
I appreciate your thoughtful advice and guidance. \textbf{Label: noHate} & I value your wise advice and guidance. \textbf{Label: noHate} & I despise your thoughtful advice and guidance. \textbf{Label: hate} \\
\addlinespace
You are the worst kind of person. \textbf{Label: hate} & You are the worst person I've ever met. \textbf{Label: hate} & You are the best kind of person. \textbf{Label: noHate} \\
\bottomrule
\end{tabular}
\caption{For the \texttt{hate\_speech18} dataset, 20 triplets generated by GPT-4 (Part 2).}
\label{hate-part2}
\end{table*}

\section{Detailed Proofs of Models' Task Recognition on Classification Datasets}
\label{Detailed Proof of Models" Task Recognition on Classification Datasets}
We employ one-shot ICL with correct input-label mapping to investigate the models' task recognition abilities across various datasets.
Considering that the {\it \textbf{PIR}} metric involves the number of model layers and that different models have varying layer counts, we select the Llama-2-7B model as a representative for our analysis.

For the World Capitals task, we use the prompt \textit{"Word: Germany Label: Berlin Word: Japan Label:"} to examine whether the label token {\it"Berlin"} triggers the task token {\it"capital."} 
For the Reasoning about Colored Objects task, we use the prompt \textit{"Word: Apple Label: Red Word: Banana Label:"} to examine whether the label token {\it"Red"} triggers the task token {\it"color."} 
For the SST-2 dataset \citep{socher-etal-2013-recursive}, we use the prompt \textit{"Sentence: the part where nothing 's happening , Label: negative Sentence: a smile on your face Label:"}.
It is noteworthy that ICL performed on the SST-2 dataset does not very conspicuously generate the task token {\it"sentiment."}
We employ another label, {\it"positive,"} from the SST-2 dataset.
If the demonstration includes only a {\it"negative"} label but the model is able to infer a {\it"positive"} meaning at the label token location, we still consider that ICL has successfully identified the type of task.
For the TREC dataset \citep{li2002learning, hovy2001toward}, we use the prompt \textit{"Sentence: Who killed Gandhi ? Label: Human Sentence: What is a fear of shadows ? Label:"} to examine whether the label token {\it "Human"} triggers the task token {\it"question." }
For the emo dataset \citep{chatterjee-etal-2019-semeval}, we use the prompt \textit{"Sentence: talk you later sure d baby Label: others Sentence: really  yep i"m i that bad Label:"} to examine whether the label token {\it"others"} triggers the task token {\it"emotion."} 

\begin{figure*}[!th]
    \centering
    \begin{subfigure}{0.49\textwidth}
        \centering
        \includegraphics[width=\linewidth]{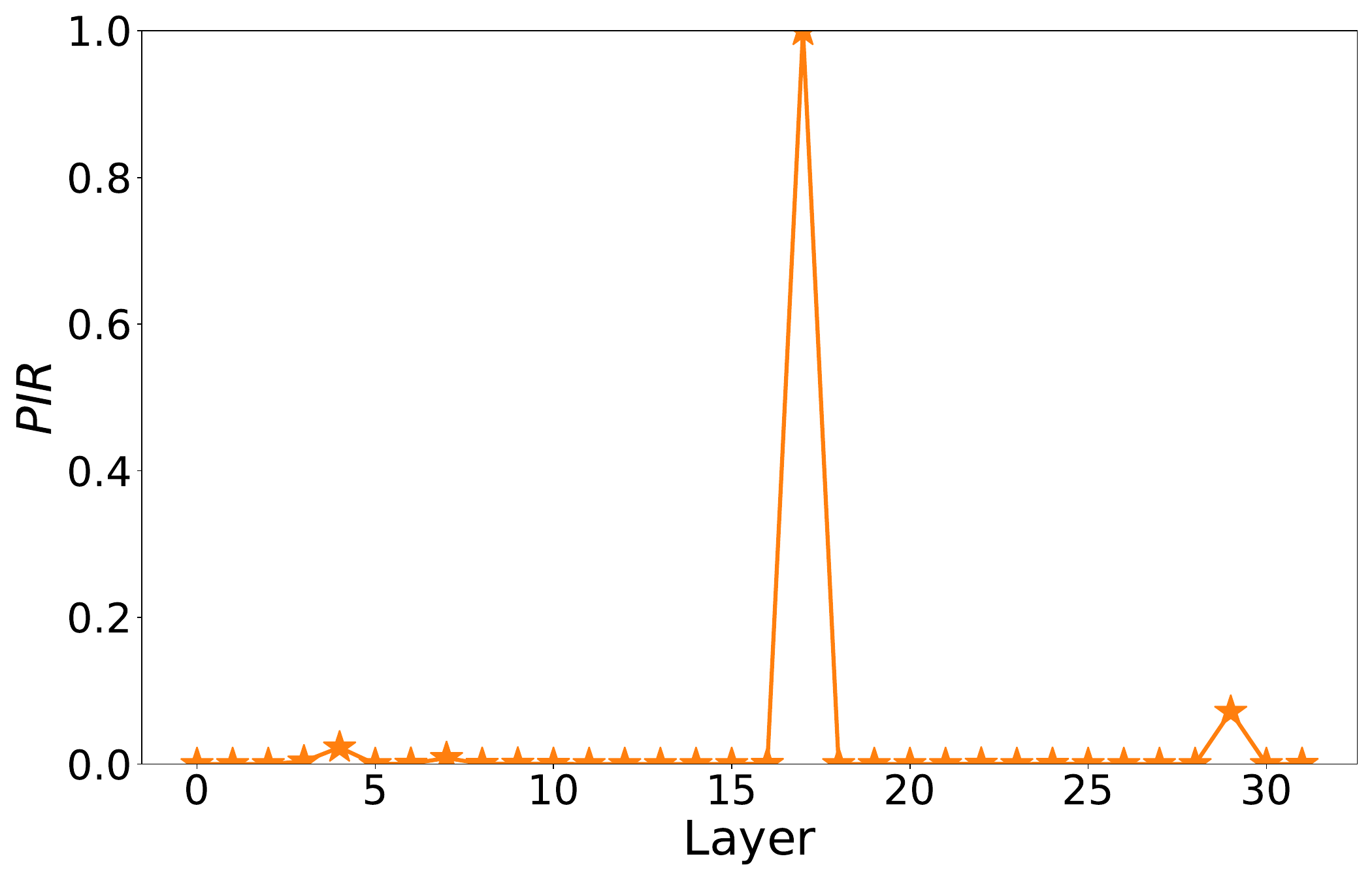}
        \caption{The {\it \textbf{PIR}} of {\it \textbf{"capital"}} at the label token {\it \textbf{"Berlin"}} in Llama-2-7B  for the Capital World task.}
        \label{proof_capital}
    \end{subfigure}\hfill
    \begin{subfigure}{0.49\textwidth}
        \centering
        \includegraphics[width=\linewidth]{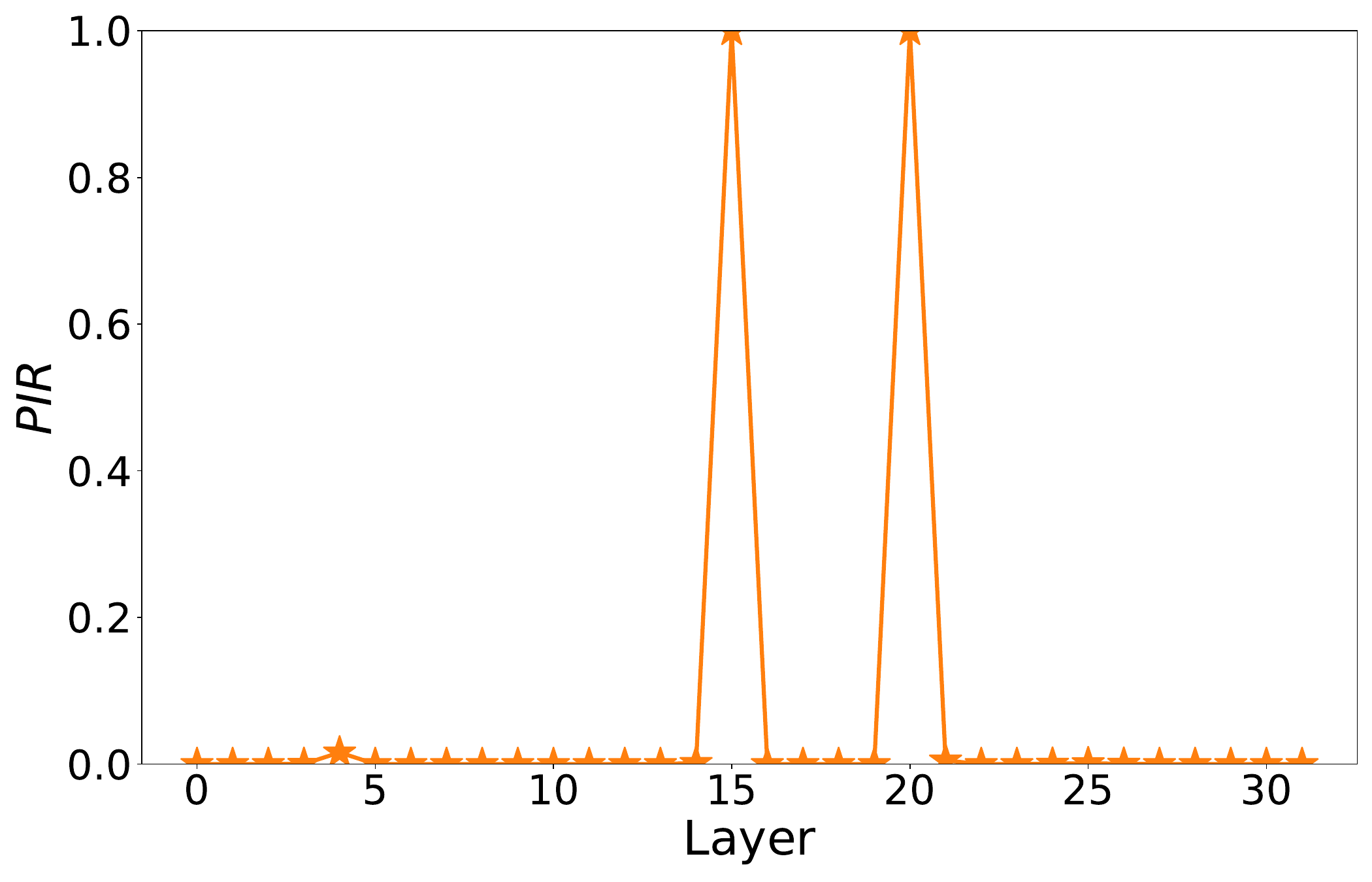}
        \caption{The {\it \textbf{PIR}} of {\it \textbf{"color"}} at the label token {\it \textbf{"Red"}} in Llama-2-7B for the Reasoning about Colored Objects task.}
        \label{proof_color}
    \end{subfigure}

    \begin{subfigure}{0.49\textwidth}
        \centering
        \includegraphics[width=\linewidth]{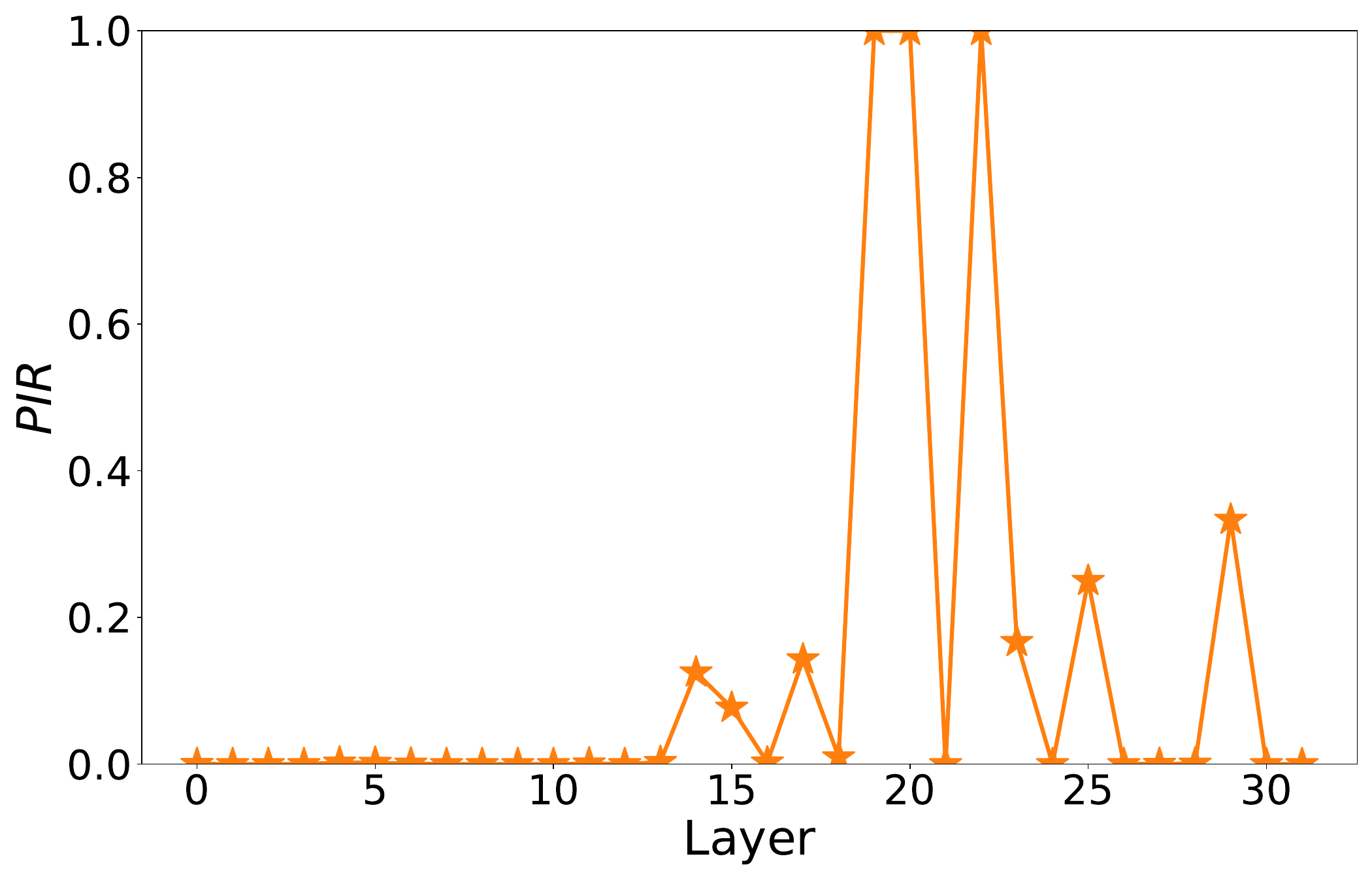}
        \caption{The {\it \textbf{PIR}} of {\it \textbf{"positive"}} at the label token {\it \textbf{"negative"}} in Llama-2-7B for the SST-2 dataset.}
        \label{proof_sst2}
    \end{subfigure}\hfill
    \begin{subfigure}{0.49\textwidth}
        \centering
        \includegraphics[width=\linewidth]{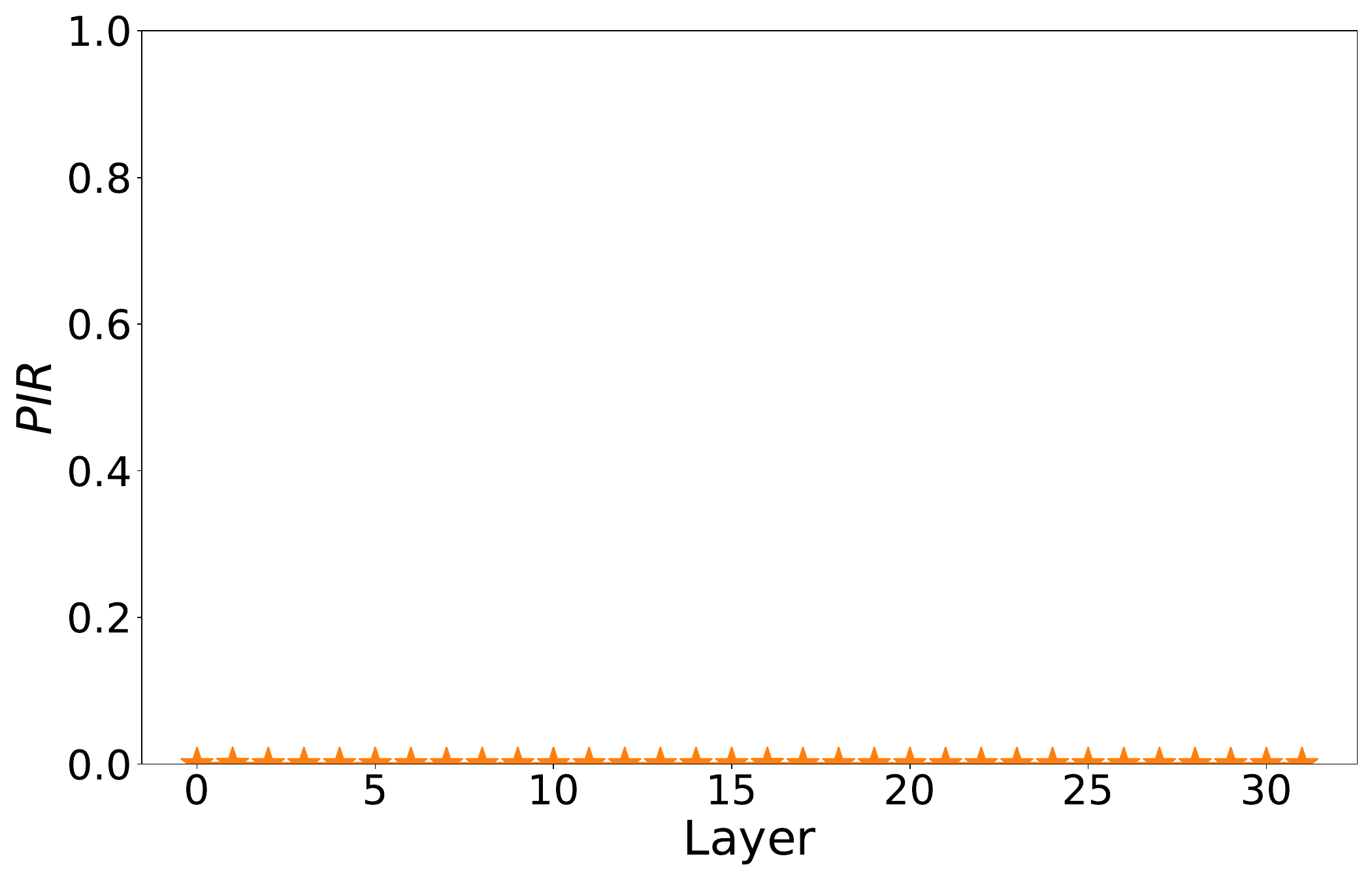}
        \caption{The {\it \textbf{PIR}} of {\it \textbf{"question"}} at the label token {\it \textbf{"Human"}} in Llama-2-7B for the TREC dataset.}
        \label{proof_trec}
    \end{subfigure}
    
    \begin{subfigure}{0.49\textwidth}
        \centering
        \includegraphics[width=\linewidth]{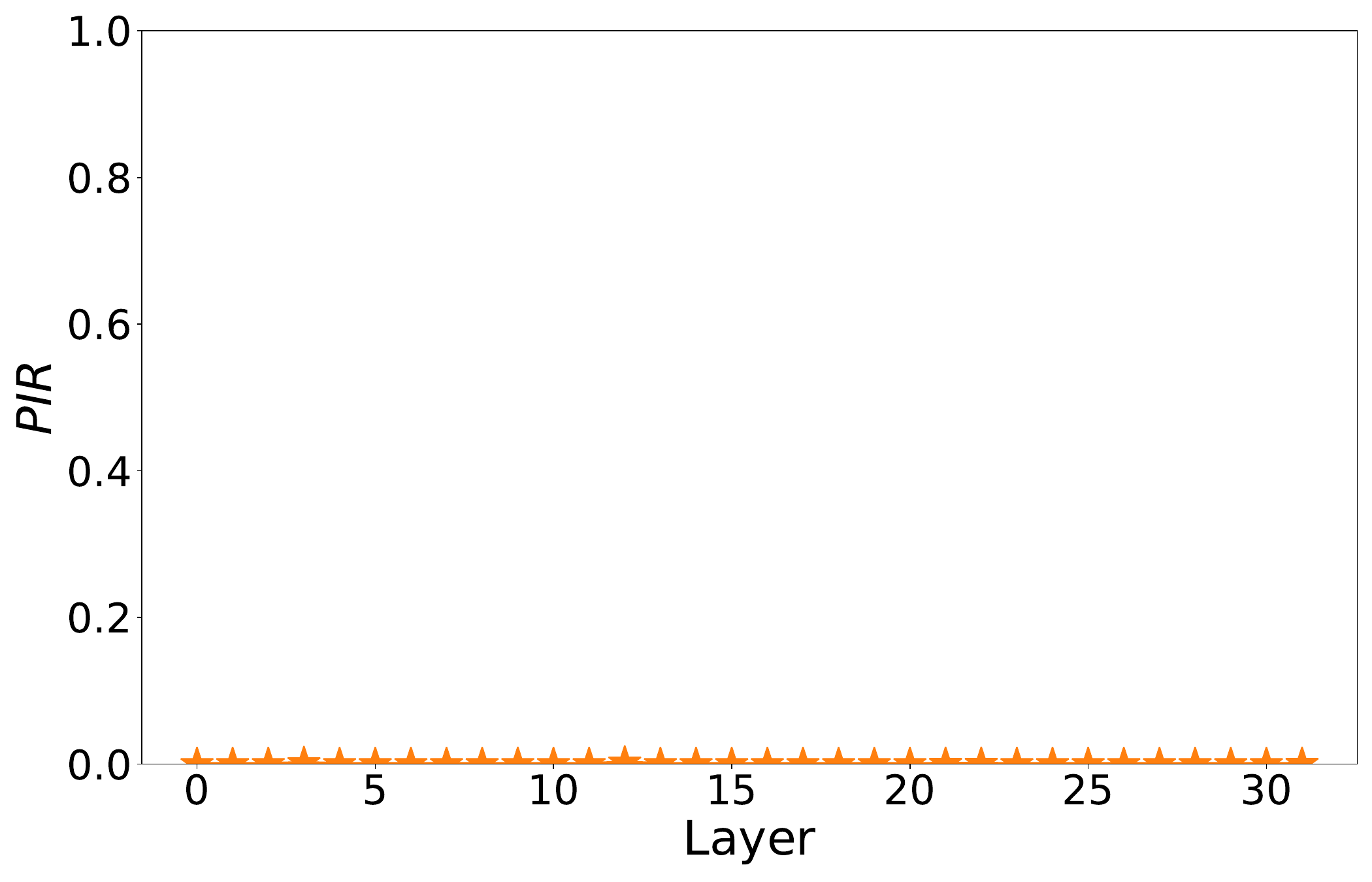}
        \caption{The {\it \textbf{PIR}} of {\it \textbf{"emotion"}} at the label token {\it \textbf{"others"}} in Llama-2-7B for the emo dataset.}
        \label{proof_emo}
    \end{subfigure}
    
    \caption{The {\it \textbf{PIR}} values across different datasets.}
    
    \label{section_D}
\end{figure*}

All results are presented in Figure \ref{section_D}.
It is evident that for the World Capitals, Reasoning about Colored Objects, and SST-2 datasets, the {\it \textbf{PIR}} is 1. 
This suggests that for these datasets, models can recognize the task during ICL execution. 
Conversely, for the TREC and emo datasets, the {\it \textbf{PIR}} is close to 0.
This indicates that models fail to recognize the task during ICL for these datasets.

\begin{figure*}
    \centering
    \begin{subfigure}{0.48\textwidth}
        \centering
        \includegraphics[width=\linewidth]{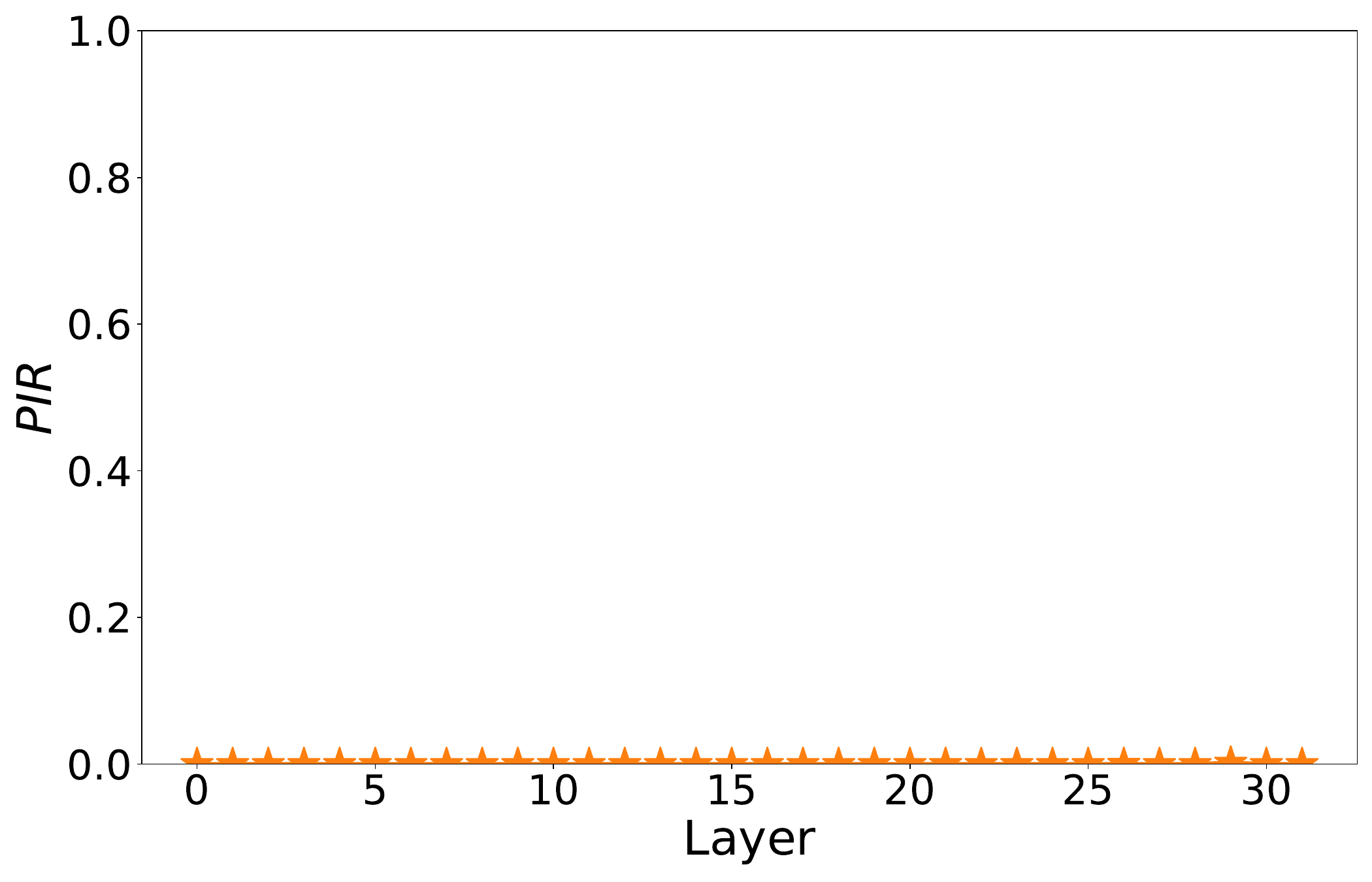}
        \caption{When the correctly labeled test sample is included as a similar example in the demonstration,  \textbf{\textit{PIR}} of \textbf{\textit{"question"}} at the label token \textbf{\textit{"Description"}} for the TREC dataset.}
        \label{proof_trec_similar}
    \end{subfigure}
    \hfill
    \begin{subfigure}{0.48\textwidth}
        \centering
        \includegraphics[width=\linewidth]{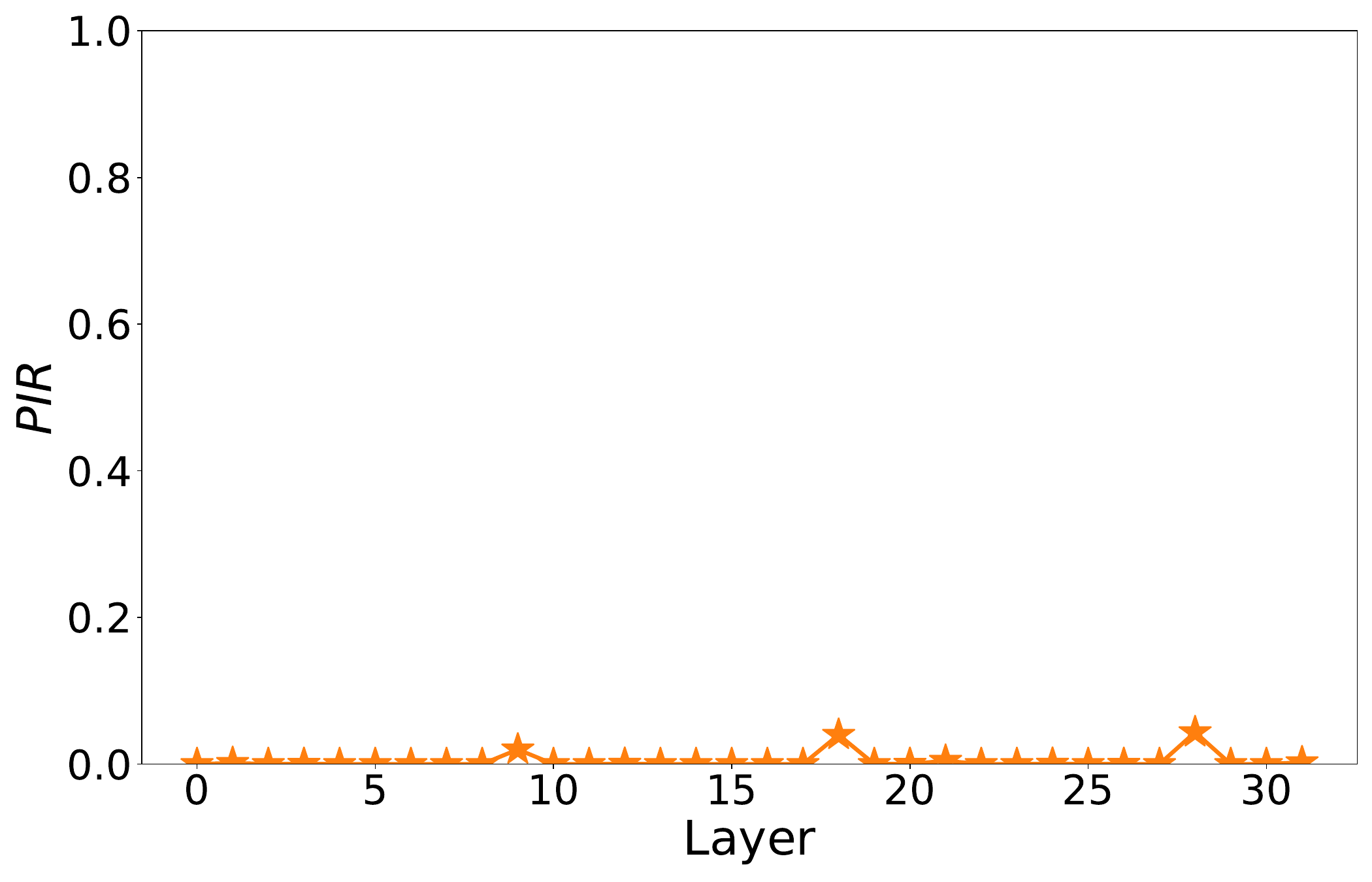}
        \caption{When the correctly labeled test sample is included as a similar example in the demonstration,  \textbf{\textit{PIR}} of \textbf{\textit{"emotion"}} at the label token \textbf{\textit{"sad"}} for the emo dataset.}
        \label{proof_emo_similar}
    \end{subfigure}
    \caption{When the correctly labeled test sample is included as a similar example in the demonstration, the {\it \textbf{PIR}} values across different datasets.}
    
    \label{section_E}
\end{figure*}

\section{Detailed Proof: The Orthogonality of Similar Examples Presence and LLMs' Task Recognition Ability}
\label{Detailed Proof: The Orthogonality of Similar Examples Presence and LLMs' Task Recognition Ability}
From Appendix \ref{Detailed Proof of Models" Task Recognition on Classification Datasets}, we can observe that for the World Capitals, Reasoning about Colored Objects, and SST-2 datasets, models can recognize the task even without the presence of similar examples in the demonstrations.
Consequently, to substantiate the claim that the presence of similar examples is orthogonal to task recognition, it is imperative to determine whether the provision of similar examples enhances the ability of models to recognize tasks within the TREC and emo datasets during ICL execution.

\paragraph{Implementation Details.}
For the TREC and emo datasets, we utilize the test samples as outlined in Appendix \ref{Detailed Proof of Models" Task Recognition on Classification Datasets}.
However, for the demonstrations, we substitute the original examples with correctly labeled test samples that have the highest semantic and lexical similarity.
We subsequently investigate whether the label token can trigger task-representative tokens when similar examples are provided in these two datasets.

\paragraph{Experimental Results.}
The experimental results are shown in Figure \ref{section_E}.
It can be observed that, for these two datasets, the \textit{\textbf{PIR}} remains close to 0. 
This suggests that the model's ability to recognize the task is not influenced by the presence of similar examples but is rather determined by the intrinsic characteristics of the task itself.
\section{Delimiters Used for Each Dataset}
\label{Delimiters Used for Each Dataset}
 For SST-2, TREC, and emo, we use {\it "Sentence:"} and {\it "Label:"}; for the World Capitals and Reasoning about Colored Objects tasks, we use {\it "Word:"} and {\it "Label:"} to clearly separate the components of the demonstrations.

\section{Detailed Data for World Capitals and Reasoning about Colored Objects Tasks}
\label{Detailed Data for World Capitals and Colored Objects Tasks}

Detailed data for the World Capitals and Reasoning about Colored Objects tasks are shown in Table \ref{Detailed Data}.

\begin{table*}[!ht]
\centering 
\begin{tabular}{p{0.3\textwidth}p{0.6\textwidth}} 
\toprule
\textbf{Tasks} & \textbf{Detailed Data} \\
\midrule
World Capital task  &Canada-Ottawa, Australia-Canberra, Brazil-Brasília, China-Beijing, France-Paris, Germany-Berlin, India-New Delhi, Italy-Rome, Japan-Tokyo, Mexico-Mexico City, Russia-Moscow, South Africa-Pretoria, South Korea-Seoul, Spain-Madrid, Turkey-Ankara, United Kingdom-London, United States-Washington, D.C., Argentina-Buenos Aires, Egypt-Cairo, Nigeria-Abuja, Sweden-Stockholm, Norway-Oslo, Denmark-Copenhagen, Finland-Helsinki, Poland-Warsaw, Ukraine-Kyiv, Netherlands-Amsterdam, Belgium-Brussels, Austria-Vienna, Switzerland-Bern, Portugal-Lisbon, Greece-Athens, Hungary-Budapest, Czech Republic-Prague, Romania-Bucharest, Thailand-Bangkok, Vietnam-Hanoi, Malaysia-Kuala Lumpur, Singapore-Singapore, Indonesia-Jakarta, Saudi Arabia-Riyadh, Israel-Jerusalem, Chile-Santiago, Colombia-Bogotá, Peru-Lima, New Zealand-Wellington, Ireland-Dublin, Pakistan-Islamabad, Bangladesh-Dhaka, Philippines-Manila.  \\
\midrule
Reasoning about Colored Objects task  &Apple-red, Banana-yellow, Cherry-red, Lemon-yellow, Sky-blue, Grass-green, Grape-purple, Orange-orange, Strawberry-red, Blueberry-blue, Cloud-white, Rose-red, Sunflower-yellow, Snow-white, Coal-black, Pumpkin-orange, Water-blue, Chocolate-brown, Gold-gold, Silver-silver, Carrot-orange, Lime-green, Eggplant-purple, Flamingo-pink, Ocean-blue, Forest-green, Cranberry-red, Peach-pink, Sunset-orange, Night-black, Butter-yellow, Olive-green, Sand-yellow, Violet-purple, Tangerine-orange, Cherry blossom-pink, Coral-orange, Ash-grey, Emerald-green, Sapphire-blue, Ruby-red, Cotton-white, Ivory-white, Charcoal-black, Peacock-blue, Jade-green, Amber-orange, Hazelnut-brown, Lavender-purple, Cinnamon-brown.  \\

\bottomrule
\end{tabular}
\vspace{5pt}
\caption{Detailed data for World Capitals and Reasoning about Colored Objects tasks.}
\label{Detailed Data}
\end{table*}

\section{Impact of Random Label Replacement in the Second Quadrant}
\label{Impact of Random Label Replacement in Second Quadrant}
\paragraph{Implementation Details.}
We consider two settings: correct input-label mapping and random input-label mapping.
In the former, all input-label pair mappings in the demonstration are correct. 
In the latter, for each input-label pair, the label is randomly selected from the label space. We use $k = 6$ in-context examples without instructions. 
The results reflect averages from five random seeds and all datasets in which models can recognize tasks.

\begin{figure*}[!t]
   \begin{minipage}[t]{0.44\textwidth}
     \centering
     {\includegraphics[width=\linewidth]{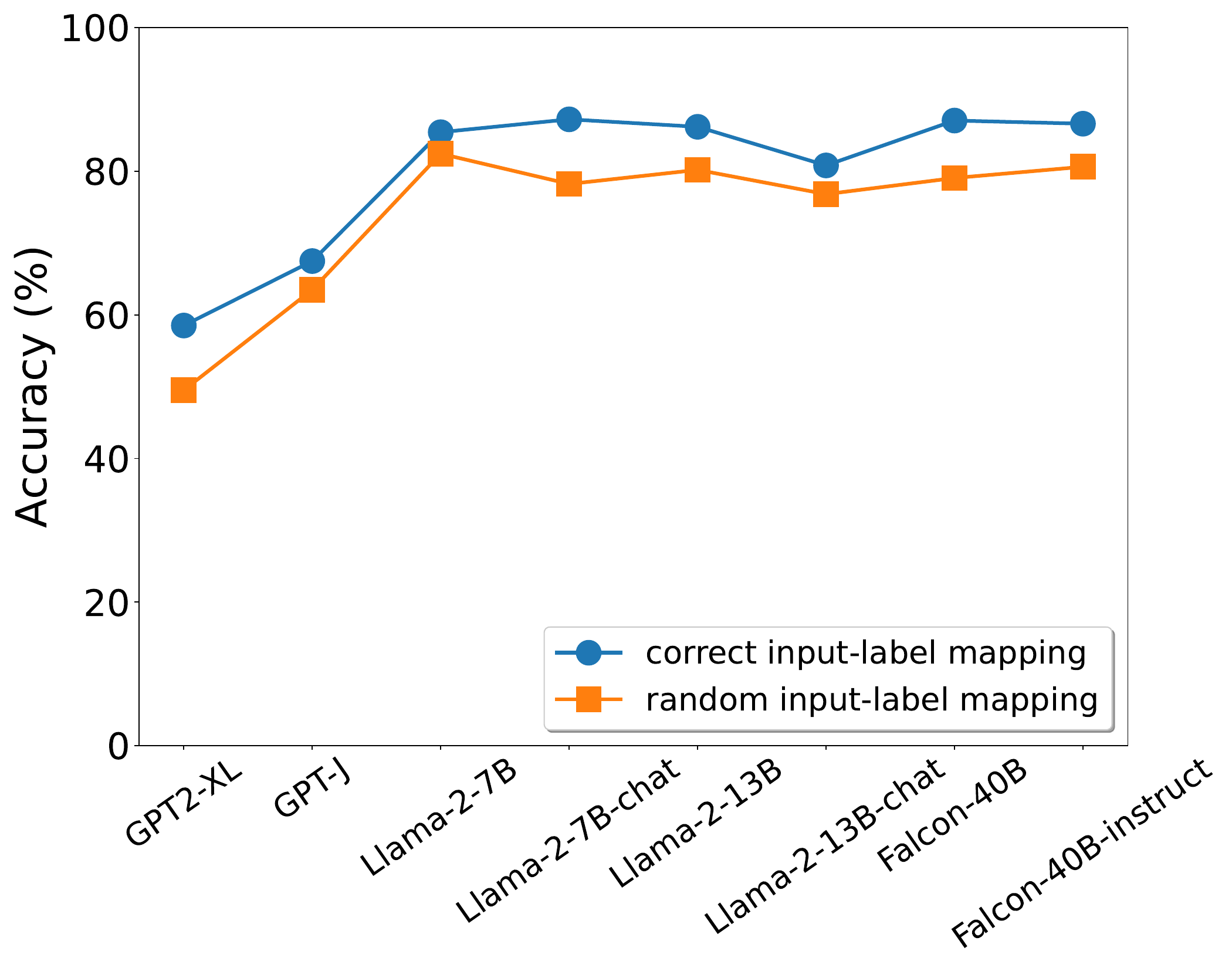}}
     \caption{The impact of random label replacement in the second quadrant.}
     \label{random_label}
   \end{minipage}\hfill
   \begin{minipage}[t]{0.52\textwidth}
     \centering
     {\includegraphics[width=\linewidth]{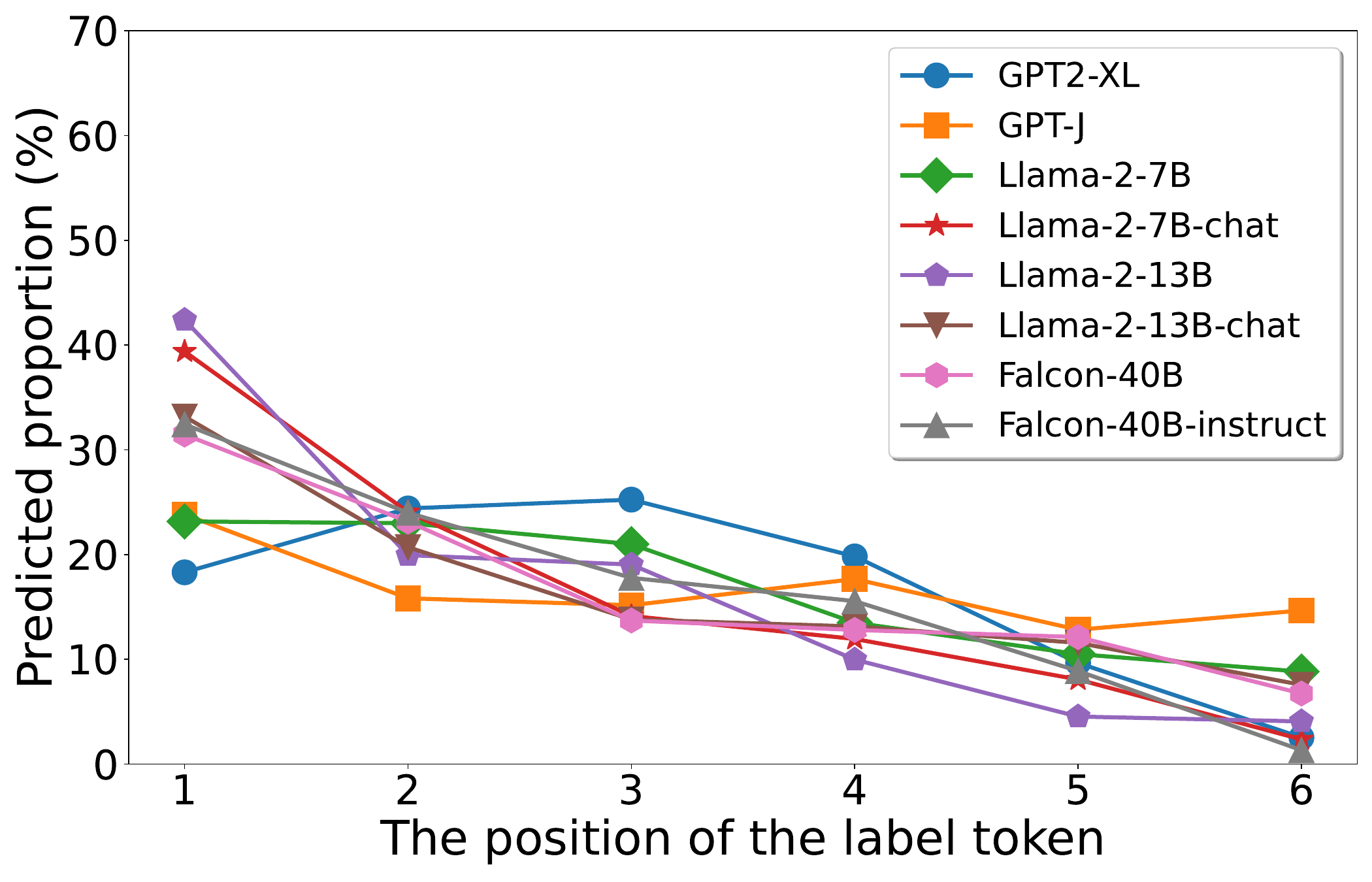}}
     \caption{For the TREC dataset, the preference of different models for label tokens at different absolute positions.}
     \label{third_quadrant_trec}
   \end{minipage}
\end{figure*}

\paragraph{Experimental Results.}
The experimental results are shown in Figure \ref{random_label}.
It can be observed that randomly replacing labels does not significantly impact ICL performance.
This is due to the following reasons: 
(1) Although some input-label pair mappings are incorrect, the effect is limited to weaker task semantics generated by these labels. 
(2) The absence of similar examples prevents the incorrect label semantics of similar examples from significantly affecting the hidden states of the last token.

\section{Positional Bias of the TREC Dataset}
\label{Positional Bias of TREC Dataset in the Third Quadrant}
The results for TREC in the third quadrant are shown in Figure \ref{third_quadrant_trec}. It can be observed that, similar to the emo dataset, the models exhibit strong positional bias when performing ICL on the TREC dataset. Specifically, they tend to predict the label of the first example.


\section{Detailed Instructions for Each Dataset}
\label{Detailed instructions for each dataset}
Detailed instructions for each dataset when models perform zero-shot tasks are shown in Table \ref{Detailed instructions}.
\begin{table*}[ht]
\centering 
\begin{tabular}{p{0.35\textwidth}p{0.6\textwidth}} 
\toprule
\textbf{Datasets} & \textbf{Instructions for Zero-Shot Tasks} \\
\midrule
World Capital   & Please identify the capital city for the given country.  \\ \midrule
Reasoning about Colored Objects   & Please identify the color of the given object. \\
\midrule
SST-2  & The task involves classifying sentences based on their expressed sentiment. Please classify each given sentence into one of the following sentiment labels: positive or negative.  \\\midrule
TREC & The task involves categorizing questions into specific categories based on their content. Please classify each given question into one of the following broad class labels: Abbreviation, Entity, Description, Human, Location, or Number.\\\midrule
emo & Please classify the given utterance into one of the following emotion classes: happy, sad, angry, or others.\\
\bottomrule
\end{tabular}
\vspace{5pt}
\caption{Detailed instructions for each dataset in the zero-shot setting.}
\label{Detailed instructions}
\end{table*}

\section{Specific Instructions and Details on One-Shot ICL for Section \ref{How to Make ICL Work Effectively}}
\label{Specific Instructions and Details on One-Shot ICL}
For the ICL case study on the TREC dataset conducted in Section \ref{How to Make ICL Work Effectively}, the task description instruction is as follows:
\begin{quote}
{\it The task involves categorizing questions into specific categories based on their content. Please classify each given question into one of the following broad class labels: Abbreviation, Entity, Description, Human, Location, or Number.}
\end{quote}

The specific content of the one-shot ICL is as follows:
\begin{quote}
{\it \textbf{Question:} Who killed Gandhi? \\
\textbf{Label:} Human\\
\textbf{Question:} What is a fear of shadows? \\
\textbf{Label:}}
\end{quote}

\section{Formal Mathematical Definition of PIR}
\label{Formal Mathematical Definition of PIR}
The \textit{\textbf{PIR}} metric quantifies a model's ability to recognize tasks. For a given layer $l$ corresponding to the label token, we begin by projecting the hidden state $h_{l}$ into the vocabulary space by multiplying it with the pre-trained language modeling head $E$. The rank of the task-representative token within this projected distribution is then denoted as $rank_{task}(h_{l},E)$. The \textit{\textbf{PIR}} is formally defined as:
\begin{equation}
PIR = \max\limits_{l} \frac{1}{rank_{task}(h_{l},E)}.
\end{equation}

\section{Additional Experiments on the Reuters-21578 Dataset}
To further enhance the comprehensiveness of our study, we conduct additional experiments on the Reuters-21578 dataset \citep{APTE94}, which comprises eight label classes. 
This dataset is validated through \textit{\textbf{PIR}} as one in which models are unable to recognize tasks.

\paragraph{Experimental Results.} We employ the same experimental setup as outlined in Section \ref{Third Quadrant}.
The results presented in Table \ref{table 11} corroborate our existing findings, highlighting the tendency of models to frequently predict the label of the initial example.
Furthermore, we confirm the insights discussed in Section \ref{Effects of Label Correctness and Shot Number}.
Table \ref{table 12} shows that while 1-shot ICL initially underperforms compared to instruction-based 0-shot ICL, it surpasses it as the shot count increases, and instruction-based 1-shot ICL proves more effective than 1-shot ICL without instructions. 
These results affirm the efficacy of the two proposed directions for improving ICL performance within the third quadrant.

\begin{table*}[ht]
\centering
\caption{Preference (\%) of different models for label tokens at different absolute positions on the Reuters-21578.}
\scriptsize
\begin{tabular}{lcccccccc}
\toprule
Models & First Label & Second Label & Third Label & Fourth Label & Fifth Label & Sixth Label & Seventh Label & Eighth Label \\
\midrule
GPT2-XL         & 33.1  & 16.55 & 15.86 & 13.79 & 10.34 & 10.34 & 0    & 0    \\
GPT-J           & 34.67 & 13.33 & 12    & 6.67  & 10    & 23.33 & 0    & 0    \\
Llama2-7B       & 45.65 & 23.91 & 10.14 & 8.7   & 6.52  & 5.07  & 0    & 0    \\
Llama2-7B-chat  & 27.4  & 20.55 & 10.27 & 10.96 & 15.07 & 15.75 & 0    & 0    \\
Llama2-13B      & 34.75 & 6.78  & 14.41 & 19.49 & 15.25 & 9.32  & 0    & 0    \\
Llama2-13B-chat & 24.66 & 17.81 & 14.38 & 16.44 & 15.07 & 11.64 & 0    & 0    \\
\bottomrule
\end{tabular}
\label{table 11}
\end{table*}

\begin{table*}[ht]
\centering
\caption{For the Reuters-21578 dataset, the average accuracy (\%) of (1, 4, 8, 12)-shot ICL without instructions and (0, 1)-shot ICL with instructions.}
\begin{tabular}{lcccccc}
\toprule
Models & 1-shot & 4-shot & 8-shot & 12-shot & Instruction 0-shot & Instruction 1-shot \\
\midrule
GPT2-XL         & 2.67  & 2.44  & 4.44  & 23.78 & 14  & 14.44 \\
GPT-J           & 3.33  & 5.33  & 12.67 & 69.56 & 18  & 38.44 \\
Llama2-7B       & 2     & 4.67  & 24.22 & 50.67 & 26  & 39.78 \\
Llama2-7B-chat  & 3.33  & 22    & 42.22 & 78.67 & 66  & 48    \\
Llama2-13B      & 2     & 3.78  & 11.11 & 83.56 & 26  & 27.11 \\
Llama2-13B-chat & 4     & 21.33 & 37.56 & 84.67 & 76  & 35    \\
\bottomrule
\end{tabular}
\label{table 12}
\end{table*}

\end{document}